# FEWT: Frequency-Enhanced Wavelet-based Transformer for Multimodal Wheeled Bimanual Manipulation

Jiaxin Huang*, Hanyu Liu*, Yunsheng Ma, Jian Shen, Yilin Zheng, Jiayi Wen, Zhigong Song†

Jiangsu Key Laboratory of Advanced Food Manufacturing Equipment and Technology, School of Mechanical Engineering, Jiangnan University, Wuxi 214122, China.

*These authors contributed equally to this work. †Corresponding author: song_jnu@jiangnan.edu.cn.

**Abstract:** Embodied intelligence bridges the physical world and information spaces, with robots demonstrating immense potential through imitation learning algorithms. In this study, a custom-built wheeled bimanual robotic platform equipped with an exoskeleton-style teleoperation system was utilized to realize intuitive remote manipulation and the efficient collection of anthropomorphic action data. To overcome the representation mismatch between spatial visual semantics and localized high-frequency physical dynamics, we propose a lightweight frequency-aligned imitation-learning framework, termed the Frequency-Enhanced Wavelet-based Transformer (FEWT). FEWT integrates two primary modules: Frequency-Enhanced Efficient Multi-Scale Attention (FE-EMA) and Time-Series Discrete Wavelet Transform (TS-DWT) to explicitly extract and align multi-scale features, improving the compatibility between spatial visual representations and temporal-frequency recalibration. Crucially, for real-world deployment, this framework is further extended into a multimodal system by seamlessly integrating a self-developed Smart Tactile Fabric (STF) sensor into the physical end-effectors, providing local contact-stress information that complements proprioceptive and chassis-motion cues in the shared multimodal representation. Experimental evaluations demonstrate that the core FEWT architecture significantly improves the success rate over the widely used Action Chunking with Transformers baseline, particularly during the most challenging phases of simulated bimanual insertion tasks. Furthermore, in complex real-world mobile and desktop manipulation tasks, the full STF-enhanced system effectively adapts to microscopic dynamic perturbations, yielding substantial performance enhancements.

## 1. Introduction

The emergence of embodied intelligence marks a paradigm shift in artificial intelligence, transitioning from virtual computations to physical world interactions. Within this realm, robots acquire complex manipulation skills primarily through imitation learning [1]. The efficacy of such data-driven policies heavily depends on two pillars: the quality of expert demonstrations and the robustness of perceptual representations. While proprioceptive feedback provides internal state estimation, navigating and interacting with unstructured environments predominantly relies on exteroceptive visual perception. However, despite substantial advances in computer vision, extracting robust and generalizable visual representations for dynamic robotic applications remains a significant bottleneck [2], [3]. Recent studies emphasize that optimizing visual representation architectures is often more critical than merely scaling dataset sizes [3].

To enhance these architectures, spatial attention mechanisms, such as Efficient Multi-Scale Attention (EMA) [4], have been widely adopted to improve feature extraction within backbone networks like ResNet [5]. Nevertheless, standard spatial attention modules exhibit a fundamental limitation: they focus exclusively on cross-spatial visual semantics while entirely neglecting frequency-domain dynamics. Integrating the Discrete Wavelet Transform (DWT) offers a powerful solution to this limitation by enabling Multiresolution Analysis. The efficacy of Haar DWT in preserving critical high-frequency details and mitigating feature degradation has been well-documented in complex vision tasks, ranging from semantic segmentation [6] and remote sensing [7] to underwater image restoration [8]. Building upon these successful applications in computer vision, we identify a profound alignment between the mathematical properties of Haar DWT and the physical demands of robotic operations. In the specific context of contact-rich bimanual manipulation, the choice of Haar DWT is driven by distinct physical and computational requirements. Unlike global frequency methods (e.g., Fourier Transform or DCT) that lose spatial-temporal localization, Haar DWT precisely identifies the exact location and timing of high-frequency transient events, such as rigid impacts or sudden slips. Furthermore, compared to ordinary CNN down sampling that destructively blurs edge details, or computationally heavy Vision Transformers (ViTs) that require massive datasets to implicitly learn feature separation, Haar DWT provides a highly efficient explicit inductive bias.

To enhance these visual representations, we propose the plug-and-play Frequency-Enhanced Efficient Multi-Scale Attention (FE-EMA) module. By introducing a frequency-domain enhancement mechanism via wavelet transform, FE-EMA dynamically fuses cross-spatial and frequency-domain features, effectively boosting the comprehensiveness and robustness of convolutional feature maps. However, while FE-EMA optimizes 2D visual representations, complex manipulation tasks inherently demand the extraction of multi-scale characteristics across long temporal windows. Inspired by frequency-enhanced channel attention mechanisms in time-series forecasting (e.g., FECAM [9]), we develop the Time-Series Discrete Wavelet Transform (TS-DWT) module. By substituting the Discrete Cosine Transform (DCT) with a 1D Discrete Wavelet Transform (DWT) along the feature dimension, TS-DWT enables the framework to capture sudden temporal dynamics and high-frequency feature responses more effectively. To support coordinated whole-body mobility, inertial measurement unit (IMU) data from the mobile chassis is seamlessly integrated with joint positions, enhancing the robot's physical awareness during dynamic navigation. Most crucially, to bridge the semantic gap between non-contact visual perception and physical contact dynamics, we introduce a self-developed Smart Tactile Fabric (STF) sensor seamlessly integrated onto the end-effectors. The continuous tactile data from the STF provides high-fidelity stress distribution feedback, completing a holistic multi-modal perception loop—comprising vision, proprioception, and touch—within our imitation learning system. This end-to-end framework is fully deployed on our custom-built wheeled bimanual robotic platform, which leverages an exoskeleton-style teleoperation system to ensure intuitive human-in-the-loop interactions and precise motion mapping during expert data collection (see **Fig. 1**). In our experimental analysis, Gradient-weighted Class Activation Mapping (Grad-CAM [10]) is utilized as an auxiliary tool to qualitatively verify the spatial focus of our visual backbone. Real-world and simulation evaluations demonstrate that our integrated architecture, termed the Frequency-Enhanced Wavelet-based Transformer (FEWT), significantly outperforms the state-of-the-art Action Chunking with Transformers (ACT [11]) baseline, with the individual efficacy of the FE-EMA, TS-DWT, and tactile modalities thoroughly validated through strict ablation studies. The superiority of the FE-EMA module in feature extraction and visual representation is validated through ablation experiments in the MuJoCo [12] simulation environment. Additionally, the effectiveness of the TS-DWT network in optimizing the Transformer [13] architecture is demonstrated, as shown in Ablation Experiment **Table 2**.

The main contributions of this paper are strictly focused on the algorithmic framework and multimodal perception integration, summarized as follows:

**(1) Frequency-Aligned Representation Learning:** We propose the Frequency-Enhanced Wavelet-based Transformer (FEWT) for contact-rich bimanual manipulation. FE-EMA introduces multiresolution frequency enhancement into spatial visual features, while TS-DWT recalibrates the fused Transformer memory before action decoding. This hierarchical design mitigates the representation mismatch between spatial semantics and high-frequency physical dynamics, enabling more effective spatial–temporal–frequency complementarity. **(2) Multimodal Physical Perception:** We develop a low-cost, flexible Smart Tactile Fabric sensor with a 15×15 sensing array and integrate its 225-dimensional contact information with vision, proprioception, and IMU feedback. The resulting multimodal representation combines global platform-motion cues with local contact-state information, improving robustness to motion disturbances, contact variations, and precision-demanding bimanual manipulation.

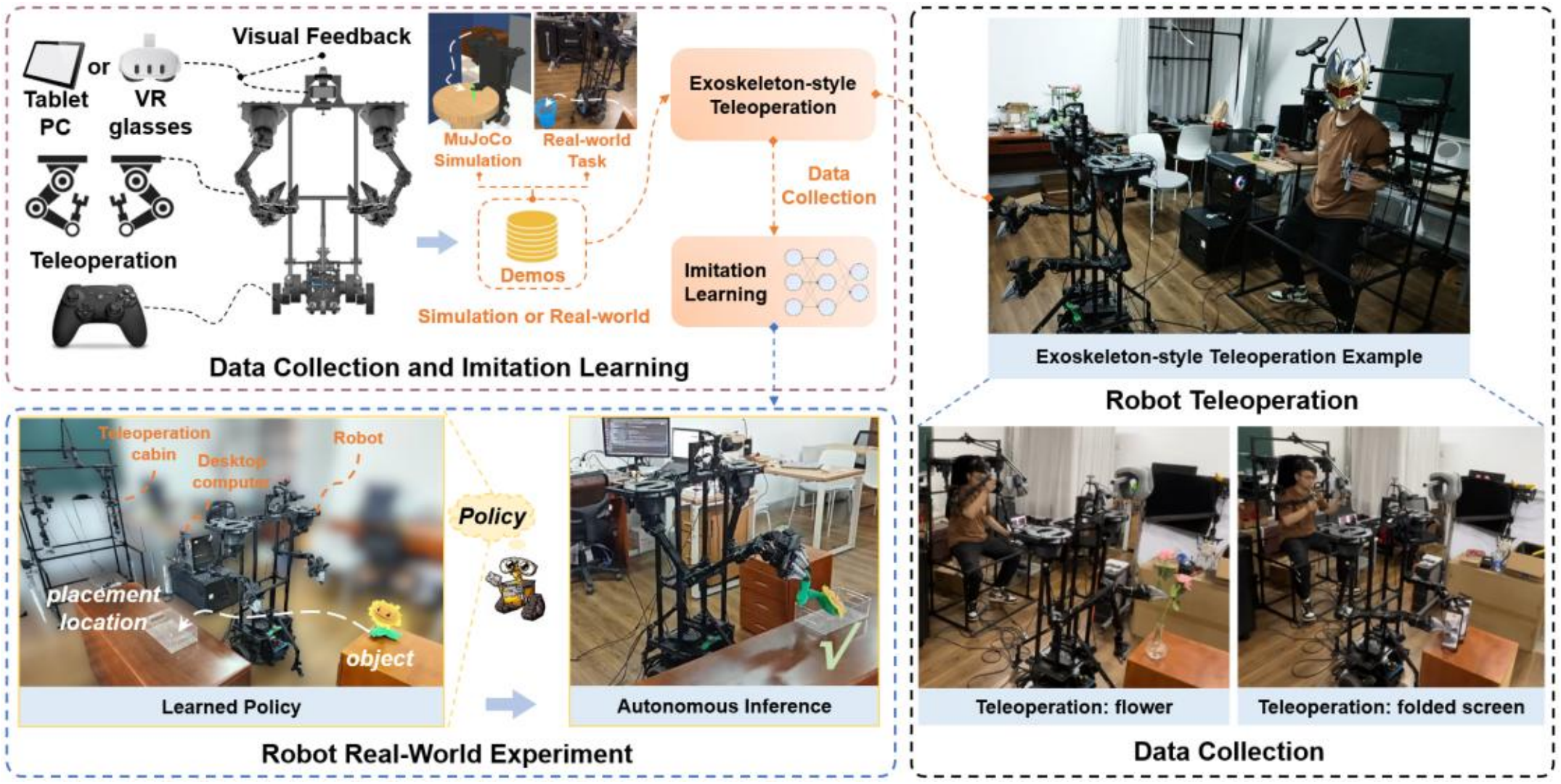

**Fig. 1.** We have developed a custom-built wheeled bimanual robotic platform. The robot learns complex bimanual tasks via imitation learning from exoskeleton-style teleoperation data, such as hand-to-hand doll transfer and box placement.

## 2. Related Work

### 2.1. Robot Teleoperation Data Collection

The datasets for imitation learning primarily consist of expert demonstrations generated through teleoperation. Teleoperation enables real-time cross-space manipulation by incorporating humans into the control loop, where humans make decisions and robots perform physical actions [14], [15]. Currently, mainstream teleoperation schemes include joint mapping [11], [16], motion retargeting [17], [18], end position control [19], and VR control [20], [21], [22], and others.

The ALOHA [11], [16] adopts a joint mapping technique, which enables teleoperation by directly mapping the joint positions of the master arm to those of the slave arm. Exoskeleton equipment improved manipulation accuracy and user experience through real-time mapping and joint feedback information [23], [24]. For example, TABLIS [24], a seat-type whole-body exoskeleton cockpit developed by the University of Tokyo, enables bilateral teleoperation of humanoid robots to overcome the space limitation of movement. The AirExo [25] exoskeleton equipment developed by Shanghai Jiao Tong University is used for data collection in imitation learning of robotic dual-arm systems, supporting joint-level teleoperation as well as learning of whole-arm manipulation in the wild. The ACE [26] teleoperation system, developed by the University of California, San Diego, is a low-cost, cross-platform visual-exoskeletons teleoperation system. This exoskeleton supports high-precision teleoperation and aids imitation learning across various platforms.

### 2.2. Robot Skill Learning

Current robotic autonomous learning heavily relies on end-to-end skill learning algorithms [27]. Imitation learning is an effective approach for robots to acquire dexterous skills through expert demonstrations [28], [29]. Behavioral Cloning (BC) [30] serves as the foundation by directly mapping expert observations to actions. Mainstream end-to-end frameworks primarily follow three routes: First, the Transformer-based [13] Google RT series (RT-1 [31], RT-2 [32], and RT-X [33]) enables multimodal task execution and open-world generalization. Second, Diffusion Policy [34] utilizes Diffusion Models [35] to fit complex multimodal action distributions. Recent advancements in this route include 3D Diffusion Policy [36] and KStar Diffuser [37], with the latter introducing a Spatial-Temporal Graph Diffusion Policy combined with kinematic regularization to achieve state-of-the-art bimanual coordination. However, the high computational cost of diffusion methods limits real-time inference. The third route is the action chunking strategy represented by ACT (Action Chunking with Transformers) [11], [16], which has become a benchmark baseline. The ACT policy is widely deployed on the ALOHA hardware platform, which includes Static ALOHA [11] for desktop tasks and Mobile ALOHA [16] for dynamic whole-body manipulation. ACT autoregressively predicts 14-DoF target joint positions at 50 Hz using CNNs and CVAEs. To optimize this imitation learning framework, Shi et al. [38] proposed Automated Waypoint Extraction (AWE), utilizing dynamic programming to decompose demonstration trajectories into minimal optimal waypoints, thereby reducing decision horizons and mitigating error accumulation. In contact-rich manipulation, fusing multimodal tactile perception is critical for capturing microscopic physical dynamics. Based on Static ALOHA, Xu et al. proposed UniT [39] for efficient tactile representation learning via VQGAN. For high-precision bimanual coordination, Huang et al. introduced 3D-ViTac [40] to fuse dense tactile and visual data into a unified 3D space. This was later extended by VT-Refine [41] for precision assembly and Touch in the Wild [42] for unstructured data collection. Additionally, Gu et al. presented TactileAloha [43], integrating GelSight sensors to capture surface textures and normal forces. To bridge the distribution gap between human demonstrations and robot execution, Kareer et al. proposed EgoMimic [44], which combines human egocentric videos and 3D hand-tracking for mixed co-training and zero-shot generalization. Building upon this, the EMMA [45] model utilizes massive egocentric data to substantially expand robotic mobile manipulation capabilities.

Finally, Vision-Language-Action (VLA) models offer strong open-world generalization [46]. OpenVLA [47] and its fine-tuning framework OpenVLA-OFT [48] provide general solutions for open platforms. Furthermore, Physical Intelligence ($\pi_0$ [49], $\pi_{0.5}$ [50], $\pi_{0.6}$ [51]) employs a hierarchical co-training architecture for complex, long-horizon household tasks. To overcome the weak interpretability of traditional VLAs, the Chain of Thought (CoT) paradigm has been introduced. ECoT [52] requires VLA models to perform explicit multi-step reasoning (e.g., task planning and state analysis) before action generation. Similarly, CoT-VLA [53] introduces a Visual CoT mechanism that autoregressively generates sub-goal images to provide highly structured intermediate reasoning for complex operations.

## 3. Methodology

### 3.1. Frequency-Enhanced Wavelet-based Transformer

In robot learning, robust visual representation underpins perception, decision-making, and planning, directly dictating task execution success. As a localized multi-scale decomposition method, wavelet transform excels at capturing details and multi-scale features. When combined with the attention mechanism, it can dynamically focus on key areas and enhance feature representation. This improves model accuracy and efficiency in target detection and other tasks. However, naively stacking spatial attention (e.g., EMA) with temporal-frequency modules causes severe feature misalignment, as standard spatial attention's focus on visual semantics inherently conflicts with the high-frequency dynamics required for contact-rich manipulation. To resolve this and achieve true spatial-temporal-frequency synergy, we propose the Frequency-Enhanced Wavelet-based Transformer (FEWT) framework (see **Fig. 2**).

Although VLA models emphasize language-conditioned generalization and high-level reasoning, diffusion policies model multimodal action distributions through iterative denoising, and visuotactile methods focus on tactile representation or cross-modal fusion, FEWT addresses a different problem: the mismatch between spatial visual semantics and localized high-frequency physical dynamics in task-specific contact-rich manipulation. FE-EMA enhances visual features through multiresolution spatial–frequency fusion, while TS-DWT recalibrates the fused multimodal Transformer memory before action decoding. Together with complementary global motion cues from the IMU and local contact feedback from tactile sensing, FEWT provides a lightweight frequency-aligned visuomotor framework.

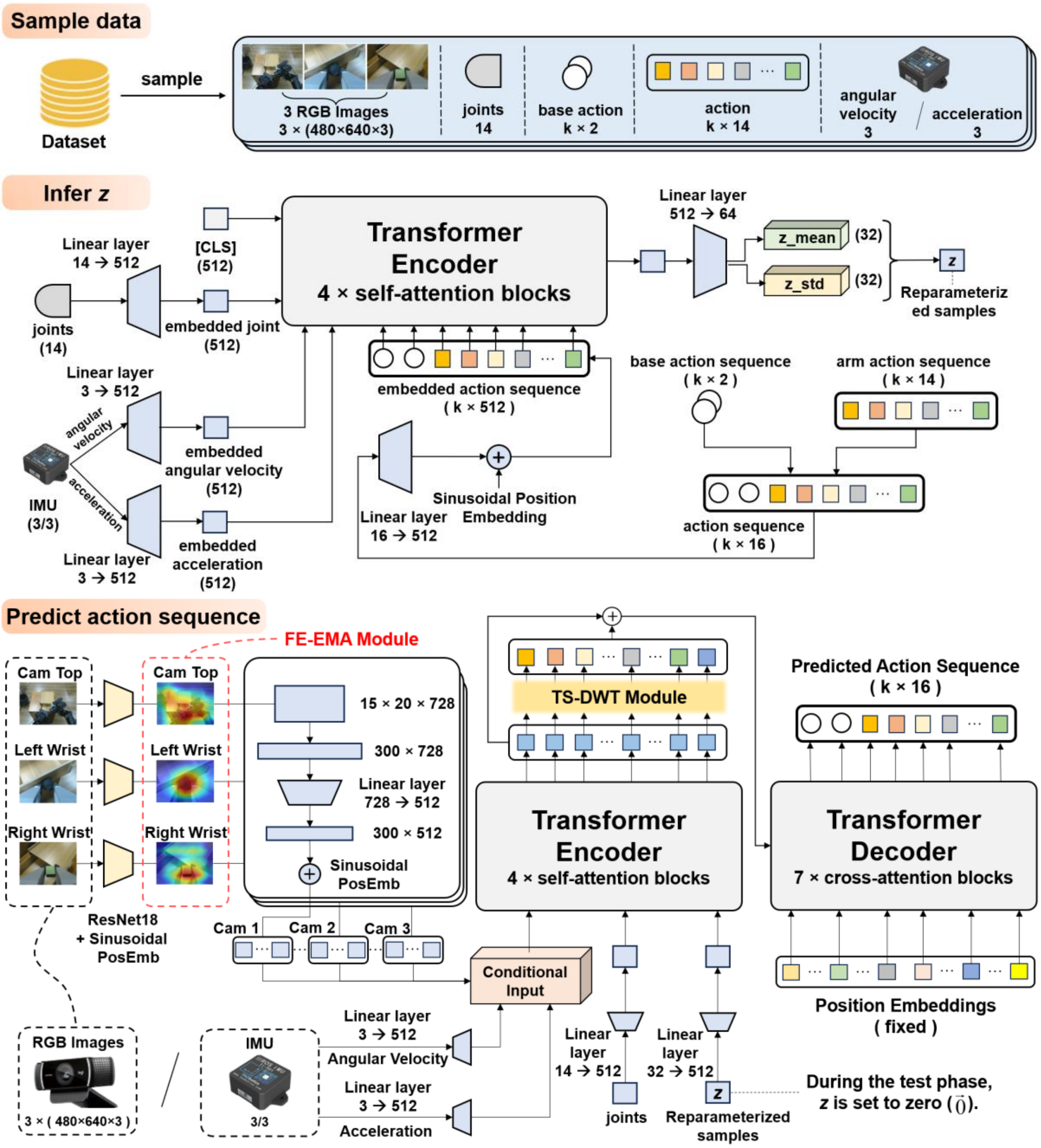

**Fig. 2.** The model framework of the Frequency-Enhanced Wavelet-based Transformer (FEWT).

Building on the original baseline model of ACT Policy, a model framework for FEWT was developed and trained as a Conditional Variational Autoencoder (CVAE), consisting of an encoder and a decoder. FEWT mainly consists of two modules, FE-EMA and TS-DWT. The encoder of the CVAE compresses the action sequences, the joint observations, and the IMU data into the "style variable" $z$. The CVAE decoder predicts a series of actions based on images from three different views. The decoder generates action sequences from images of three views, joint positions, IMU data, and $z$ using the Transformer encoder. It then predicts actions through the Transformer decoder. In the testing phase, $z$ is simply set to the mean of the prior (i.e., zero). In this case, the IMU data of the moving chassis contains the linear acceleration and angular velocity. Integrating IMU data into the model training process optimizes motion control accuracy of the mobile chassis. It also improves the performance of the robot in dynamic environments by providing precise linear acceleration and angular velocity information. In **Fig. 2**, "k" denotes the length of the target action sequence in the demonstration dataset.

### 3.2. Frequency-Enhanced Efficient Multi-Scale Attention

To overcome the inherent limitations of standard EMA—which lacks frequency awareness and triggers destructive interference during multimodal temporal decoding—FE-EMA is fundamentally redesigned to serve as a frequency-aligned spatial bridge. It implements two key architectural breakthroughs: (1) Frequency-domain Alignment & Enhancement: By incorporating wavelet transform within the spatial residual network, it ensures that the extracted visual semantics are explicitly aligned with the high-frequency physical dynamics; (2) Adaptive Fusion: dynamically adjusting the fusion ratio of the cross-spatial and frequency-domain features using the weighting parameter $\alpha$, thereby preventing the suppression of critical contact signals.

Mathematical expressions related to the discrete wavelet transform on 2D feature maps are given from the perspective of Multiresolution Analysis. Given an input feature map $X$, its wavelet transform represents the signal as **Equation (1)** by constructing a set of scaled and translated wavelet basis functions $\psi(\cdot)$ and scale functions $\phi(\cdot)$.

$$X(\mathrm{x})=\sum_{\mathrm{k}\in \mathrm{Z}^2} cA_{\mathrm{k}}\cdot\phi_{\mathrm{k}}(\mathrm{x})+\sum_{\mathrm{k}\in \mathrm{Z}^2} cH_{\mathrm{k}}\cdot\psi_{\mathrm{k}}^{\mathrm{H}}(\mathrm{x})+\sum_{\mathrm{k}\in \mathrm{Z}^2} cV_{\mathrm{k}}\cdot\psi_{\mathrm{k}}^{\mathrm{V}}(\mathrm{x})+\sum_{\mathrm{k}\in \mathrm{Z}^2} cD_{\mathrm{k}}\cdot\psi_{\mathrm{k}}^{\mathrm{D}}(\mathrm{x}) \tag{1}$$

Where $\mathrm{x} = (x, y)$ denotes the 2D spatial coordinates, and $\mathrm{k} = (k_1, k_2) \in \mathrm{Z}^2$ is the 2D spatial translation vector. $\phi_{\mathrm{k}}(\mathrm{x})$ is the 2D scaling function, while $\psi^{\mathrm{H}}(\cdot)$, $\psi^{\mathrm{V}}(\cdot)$, and $\psi^{\mathrm{D}}(\cdot)$ represent the 2D wavelet basis functions in the horizontal, vertical, and diagonal directions, respectively. Correspondingly, $cA_{\mathrm{k}}$ is the low-frequency approximation coefficient, and $cH_{\mathrm{k}}$, $cV_{\mathrm{k}}$, $cD_{\mathrm{k}}$ are the high-frequency detail coefficients for the horizontal, vertical, and diagonal directions. Z denotes the set of integers.

On a 2D feature map $X$, the wavelet transform can be implemented using a 2D filter bank. The wavelet decomposition of the feature map by rows and columns respectively yields the following coefficients: $\mathrm{DWT}(X) = (cA, cH, cV, cD)$. Where $cA$ denotes the low-frequency approximation component, capturing the overall structural information. $cH$ denotes the horizontal detail component, capturing the horizontal edge features. $cV$ denotes the vertical detail component, capturing the vertical edge features. $cD$ denotes the diagonal detail component, capturing the edge variations in the diagonal direction. The convolution kernel of Haar wavelet is defined in **Equation (2)**, where $L$ is the low-pass filter, which is used for the extraction of low-frequency components. $H$ is a high-pass filter for extracting high-frequency components.

$$L=\frac{1}{\sqrt{2}}\begin{bmatrix}1\\1\end{bmatrix},\quad H=\frac{1}{\sqrt{2}}\begin{bmatrix}1\\-1\end{bmatrix} \tag{2}$$

The convolution of a 2D feature map is divided into two steps: horizontal and vertical. Horizontal convolution uses an $L$ filter to extract the low-frequency components $cA$ and an $H$ filter to extract the detail components $cH$. Vertical convolution uses an $L$ filter to obtain the detail components $cV$ and an $H$ filter to obtain the detail components $cD$.

The final coefficients $cA$, $cH$, $cV$, $cD$ is obtained by combining the results of the horizontal and vertical convolution. The FE-EMA module decomposes the features into low-frequency and high-frequency components using the DWT. It then combines these components with ResBlock for frequency-domain feature enhancement. The goal of applying DWT on the feature map is to decompose the input feature map $X$ into coefficient matrices of different frequency bands. This decomposition extracts multi-scale and multi-frequency feature information. DWT decomposes the feature map into multi-frequency components and reduces its size. Subsequent convolutions act only on the low-dimensional features to improve computational efficiency. Meanwhile, FE-EMA module uses grouped convolution to process only the down-sampled features. This approach significantly reduces computation (FLOPs) and improves model performance and efficiency. A detailed comparison of model parameters and FLOPs is provided in supplementary material **Table S1** and **Table S2**. The network architecture of the FE-EMA module is shown in **Fig. 3**.

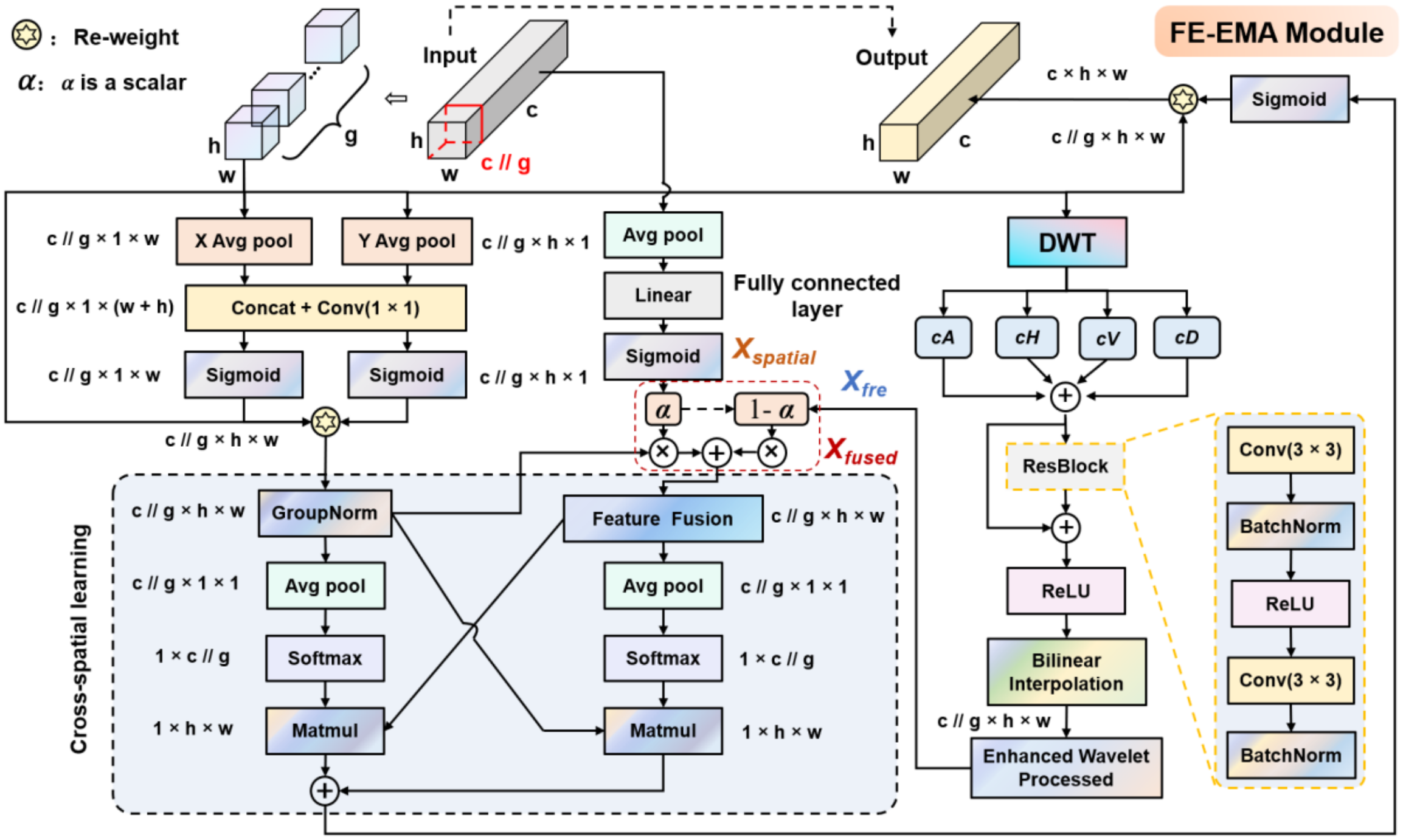

**Fig. 3.** Network architecture of the Frequency Enhanced Efficient Multi-Scale Attention (FE-EMA) module.

On a 2D feature map $X$, the wavelet transform can be implemented using a 2D filter bank. In the FE-EMA module, the high-frequency feature fusion is denoted as $F_H$ = c$H$+c$V$+c$D$. It is then summed with the low-frequency component $cA$ to obtain the enhanced feature: $X_{fre}$ = ResBlock(c$A$+$F_H$). In ResBlock, wavelet decomposition features are enhanced using 3×3 convolution and noise is suppressed by nonlinear activation to improve the capture of fine-grained information. The 1×1 branch in FE-EMA module captures local cross-channel interactions and embeds global spatial information through global average pooling.

Dynamic weights $\alpha$ are used to control the fusion of cross-spatial and frequency-domain features ($\alpha$ is a scalar). The weights $\alpha$ are generated by global average pooling and a linear layer: $\alpha = \sigma(W \cdot \mathrm{GAP}(X))$, where $\sigma(\cdot)$ is the Sigmoid function, GAP is the global average pooling, and the weight matrix $W$ is linearly transformed on this feature vector to learn the weight assignments of the features.

The spatial and frequency-domain fused feature $X_{fused}$ is shown in **Equation (3)**. Ultimately, the module generates a cross-space attention map using matrix dot product. This accurately captures pixel-level relationships and enhances feature representation.

$$X_{fused} = \alpha \cdot X_{spatial} + (1-\alpha) \cdot X_{fre} \tag{3}$$

$X_{spatial}$ in **Equation (3)** is a spatial feature generated through the attention mechanism and group normalization process. It mainly contains key regional information extracted from the input features. It is combined with the frequency-domain feature $X_{fre}$ to ensure a more comprehensive information capture of the model at multiple scales. When computing $\alpha$, global features extracted by global average pooling (Avg pool) offer an overall overview of the image. These features reflect the general nature of the input and help the model assess the importance of features in both cross-spatial and frequency-domain. Using average activation values, the model dynamically adjusts feature fusion weights and adaptively optimizes the feature fusion process in complex tasks.

### 3.3. Time-Series Discrete Wavelet Transform

The application of Discrete Wavelet Transform (DWT) in image processing is well established. However, its potential for time-series data modeling in robotic manipulation has not been fully explored. Similar to spatial images, time-series memory states contain rich multi-scale frequency features. High-frequency components often correspond to sudden dynamic changes (e.g., collision, grasping), while low-frequency components reflect stable, continuous motions. Therefore, building on the successful application of the DWT in vision tasks, this work incorporates DWT into time-series analysis to enhance the predictive model's ability to capture multi-scale representations. Unlike standard temporal forecasting modules (e.g., FECAM), imitation learning strictly requires bridging global temporal contexts with highly complex visual representations to generate continuous action sequences. Thus, we designed the Time-Series Discrete Wavelet Transform (TS-DWT) module (**Fig. 4**) as a frequency-recalibration bridge rather than an isolated filter. Inserted between the Transformer Encoder and Decoder, TS-DWT applies 1D-DWT along the feature dimension at each time step. This dynamically recalibrates frequency responses across the entire temporal window, aligning global temporal contexts with the frequency-enhanced spatial features to enable robust action chunking.

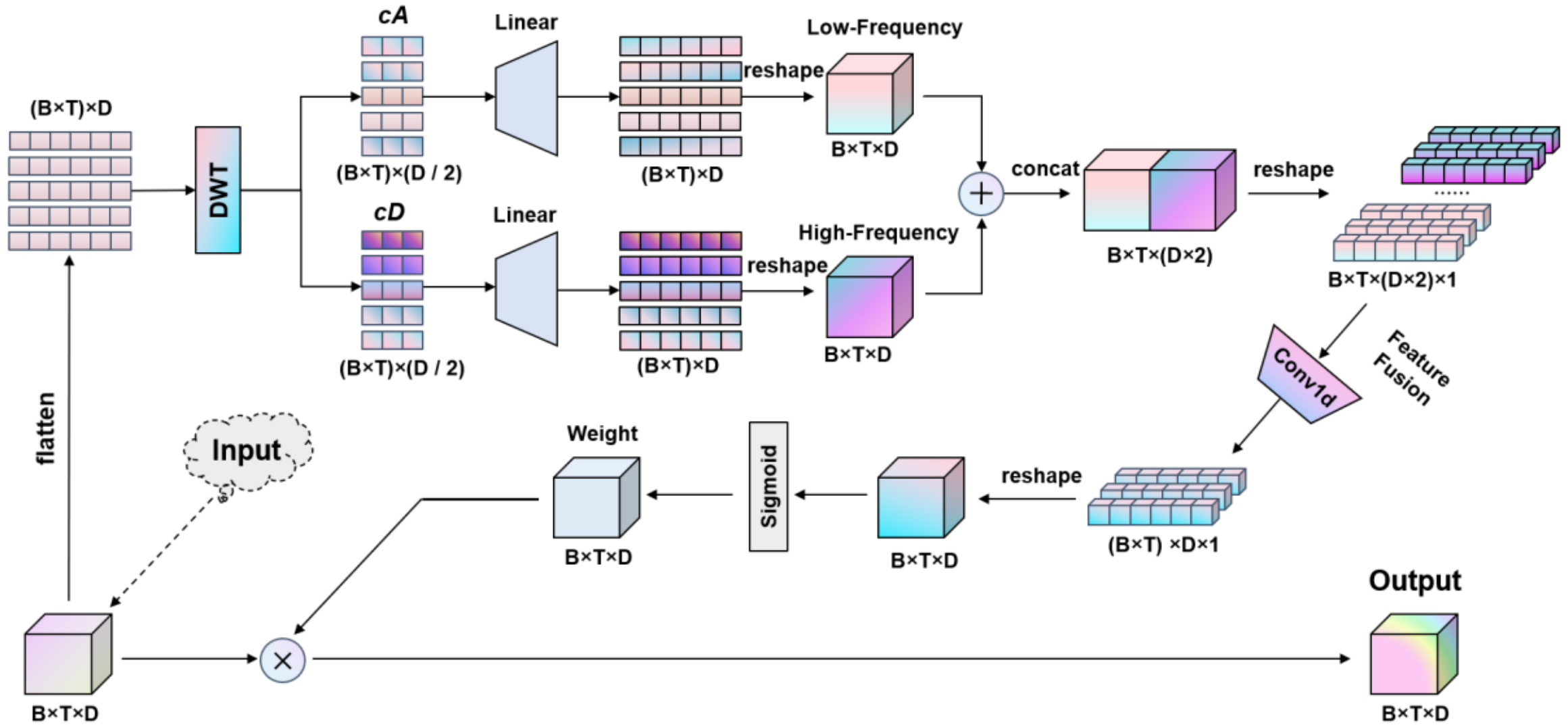

**Fig. 4.** Network architecture of the Time-Series Discrete Wavelet Transform (TS-DWT) Module.

To clarify the internal tensor operations and reshaping mechanisms, we define the input to the module as the hidden memory state $X \in R^{B\times T\times D}$ from the Transformer Encoder, where $B$ is the batch size, $T$ is the sequence length (time steps), and $D$ is the feature dimension (e.g., $d_{model} = 512$). The concise algorithmic pipeline proceeds as follows:

Flattening and Wavelet Decomposition: The input sequence $X$ is first reshaped into a 2D tensor of shape ($B\times T$, $D$) by collapsing the batch and time dimensions. A 1D Haar wavelet transform is then applied along the feature dimension $D$, decomposing it into a low-frequency approximation $cA \in R^{(B\times T)\times(D/2)}$ and high-frequency details $cD \in R^{(B\times T)\times(D/2)}$.

Independent Branch Processing: The decomposed components $cA$ and $cD$ are reshaped back to the sequence format ($B$, $T$, $D/2$) and passed through independent fully connected branches. These branches upscale the sub-bands back to the original feature dimension, yielding enhanced features $cA' \in R^{B\times T\times D}$ and $cD' \in R^{B\times T\times D}$.

Tensor Reshaping and Grouped Convolution Fusion: The processed low- and high-frequency features are concatenated along the feature axis to form a combined tensor of shape ($B$, $T$, 2D). To efficiently fuse these cross-frequency features while minimizing computational overhead (FLOPs), we employ a 1D grouped convolution (Conv1d). To align with the standard Conv1d input format, the combined tensor is deliberately reshaped into a 3D format ($B\times T$, $2D$, 1), where $B\times T$ acts as the effective batch size, $2D$ acts as the input channels, and 1 represents a dummy spatial length. The grouped Conv1d layer (kernel size 1) fuses this into ($B\times T$, $D$, 1).

Attention Generation and Residual Output: The dummy spatial dimension is squeezed, and the tensor is reshaped back to the original sequence format ($B$, $T$, $D$). Finally, frequency-domain attention weights are generated via a Sigmoid activation function and multiplied element-wise with the original input $X$ through a residual connection, yielding the final frequency-enhanced memory state for the Decoder.

## 4. Experiment and Results

### 4.1. System Platform and Data Collection

The construction of the robot (mobile dual-arm manipulation platform) mainly consists of two 7-dof (arm + gripper) robotic arms (ViperX-300) [11] , a two-wheeled differential mobility chassis (Diablo) [54], three RGB cameras (Logitech C922x webcams), and an inertial measurement unit (IMU). The top is equipped with a visual feedback screen and a camera to monitor the bimanual manipulation. We employ an inverted robotic arm as the master controller to construct an exoskeleton-style teleoperation system for collecting expert demonstration data. This setup enables high-precision manipulation and more natural teleoperation for robots. Compared to the dual-arm base link conventional layout design of Mobile ALOHA, the dual-arm inverted design of the robot is more anthropomorphic in appearance, more natural in manipulation. **Fig. 5(a)** shows the overall structural design of the robot, along with exoskeleton-style teleoperation, and provides a detailed description of the hardware on both the master and slave sides. The overall structural design and joint definition of the robot is shown in **Fig. 5(b)**. The total height (along the $z$-axis), total width (along the $y$-axis), and total length (along the $x$-axis) of the robot in the two-armed drooped state is 1317 mm, and the overall weight is about 35 kg. The structural design of the control system of the robot is shown in **Fig. 5(c)**. For data collection of the robot specific skill, it is sufficient to collect about 25 to 50 sets of expert demonstrations [16]. Remote manipulation master-slave control in data collection is achieved through Robot Operating System (ROS). By learning from the expert demonstration, it is possible to obtain actions and policies related to a specific task from the samples. The specific performance parameters of the robot are shown in **Fig. 5(d)**. Additionally, the system integrates a customized, low-cost Smart Tactile Fabric (STF) module (**Fig. S1**). Inspired by 3D-Vitac [40], this scalable e-textile utilizes a piezoresistive layer to translate contact deformations into resistance changes for real-time tactile decoding. To enable closed-loop grasping, a 15×15 STF sensing array is integrated onto the bimanual end-effectors. This flexible electronic skin provides excellent surface adaptability and high-fidelity contact perception, facilitating real-time posture correction and dexterous bimanual manipulation (**Fig. 5e**).

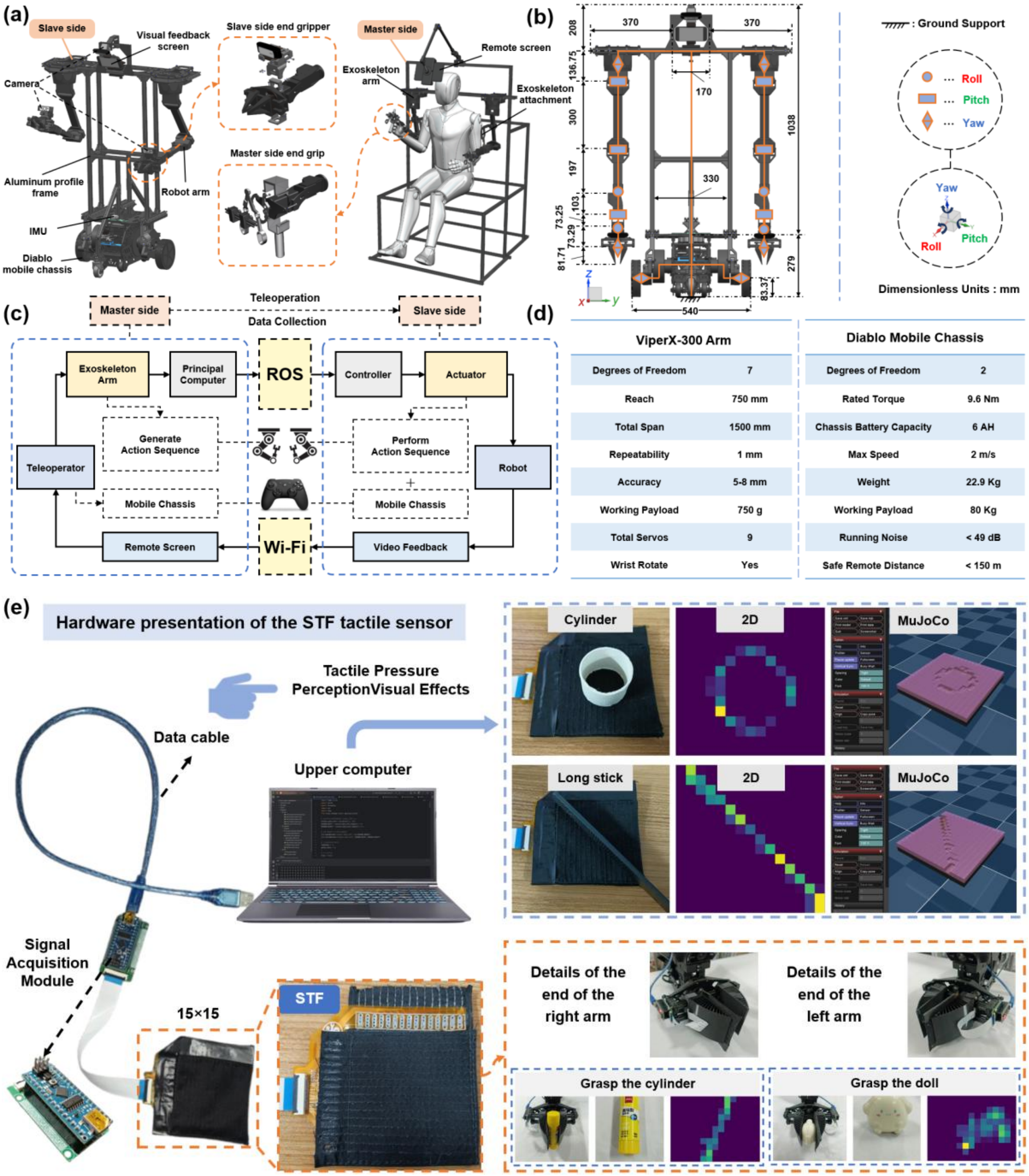

**Fig. 5.** The overall structure and hardware of the robot, along with exoskeleton-style teleoperation. (a) On the overall structural design of exoskeleton-style teleoperation. (b) Overall structural design and joint definition for robots. (c) Control system architecture for teleoperation of the robot with users. (d) Performance parameters of the robot (ViperX-300 robotic arm with Diablo mobile chassis). (e) Perceptual visualization of tactile fabric sensors and their integration into the end gripper.

By securing the exoskeleton-style master side robotic arm to the upper arm of the human operator using bandage, comfortable and natural remote manipulation can be achieved. The damping rotation device shown in **Fig. 6** allows for flexible control of the damping force by adjusting screw nut to apply positive pressure. Expert data is collected through the exoskeleton-style teleoperation system, enabling autonomous adjustment and high-precision operations without the need for gravity compensation. This approach not only effectively reduces the operator's burden during data collection but also enhances the comfort and efficiency of robotic teleoperation. The elbow joint of the master side robotic arm is equipped with an exoskeleton connector, which houses a damping rotation device to alleviate the gravity-induced sagging of the elbow joint during teleoperation. Additionally, **Figure 6** presents a comparison between our adopted exoskeleton-style teleoperation and the standard upright bimanual configuration used in Mobile ALOHA. From an ergonomic perspective, the standard upright base configuration typically forces operators to maintain a suspended or forward-reaching arm posture, which easily induces shoulder fatigue during prolonged data collection. Conversely, our inverted anthropomorphic configuration aligns with the natural downward resting posture of human upper limbs. This design allows the operator's arms to drape naturally under gravity, significantly reducing musculoskeletal strain and providing a more intuitive and relaxed teleoperation experience.

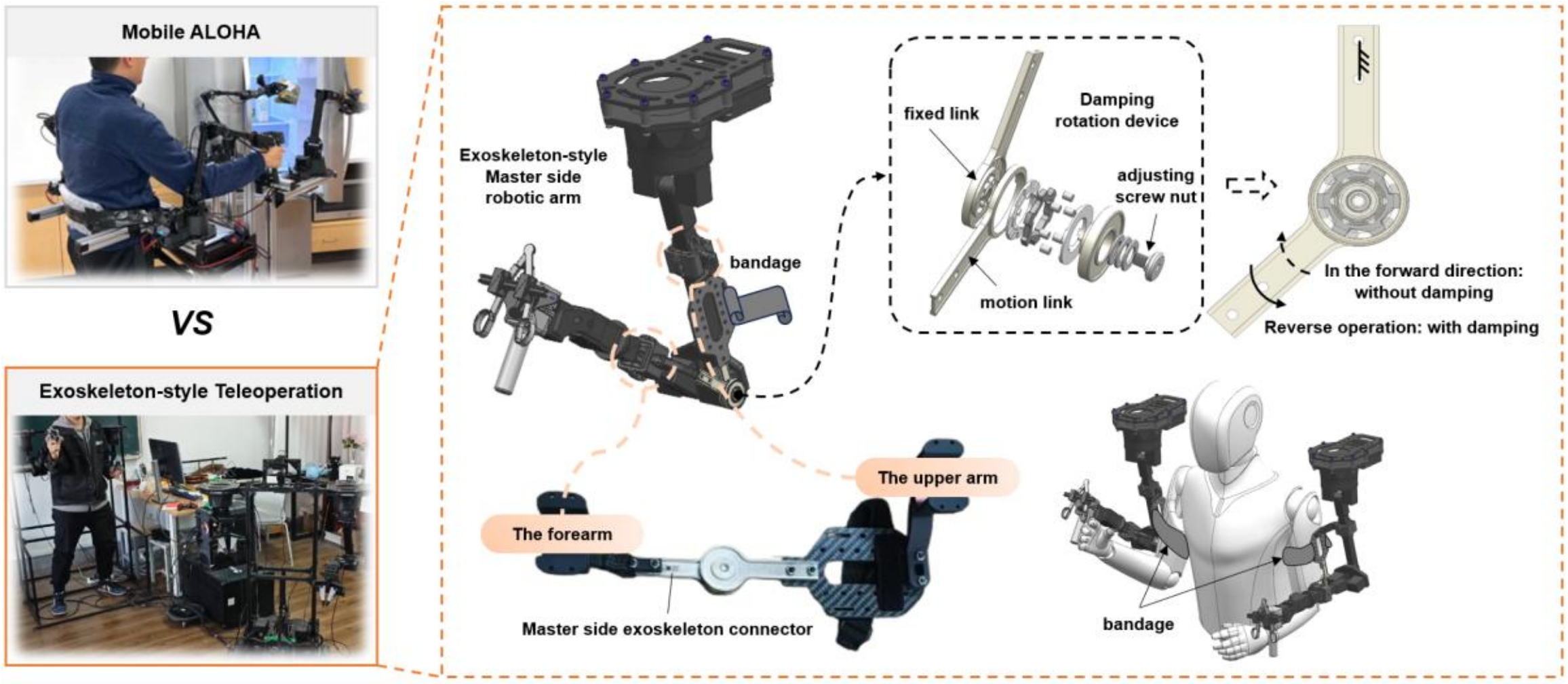

| Teleoperation methods | Direct Joint Mapping | Inverted Anthropomorphic Configuration | Upper-Arm Strapping | Passive Gravity Compensation |
|---|---|---|---|---|
| Exoskeleton-style Teleoperation (ours) | √ | √ | √ | √ |
| Mobile ALOHA | √ | × | × | × |

**Fig. 6.** Comparison of teleoperation methods and illustration of the damping rotation device in the exoskeleton connector.

## 4.2. Ablation Experiment

Based on the MuJoCo simulation environment, the model learning effect of FEWT is verified through simulation experiments. The dataset generated by the simulation task contains 50 sample sets, with all data uniformly stored in HDF5 file format. The object initialization positions for each simulation task are randomly generated within a given interval. The dataset in **Table 1** contains only the top camera view (Top Image). Length denotes the overall time step for the simulation task, with delay time DT = 0.02 seconds between each time step (control frequency of 50Hz). "Num" denotes the number of samples in the dataset, and "Total size" denotes the overall memory size of the dataset in Gigabytes (G). A summary of the specific data from the ablation experiments is given in **Table 2**. For the two simulated tasks, we report training with scripted data, with 6 seeds and 50 policy evaluations each. The experimental results show that FEWT significantly outperforms the ACT original baseline policy in terms of simulation task success rate.

Specifically, FEWT increases the success rate of the ACT baseline by up to 30 percentage points (from 32% to 62%) on the most challenging 'Insert' phase of the Bimanual Insertion simulation task. Details regarding the training of the model can be found in supplementary material **Table S3**.

**Table 1.** Summary of task-specific datasets and phases for robotic manipulation in act simulation baseline

| Dataset | Phase1 | Phase2 | Phase3 | length | num | Total size |
|---|---|---|---|---|---|---|
| *Cube Transfer* (sim) | Touched | Lifted | Transfer | 400 | 50 | 18.4G |
| *Bimanual Insertion* (sim) | Grasp | Contact | Insert | 400 | 50 | 18.4G |

**Table 2.** Comparison of success rate (%) for ACT simulation tasks in ablation experiment

| Method | *Cube Transfer* (sim) | | | *Bimanual Insertion* (sim) | | |
|---|---|---|---|---|---|---|
| | Touched | Lifted | Transfer | Grasp | Contact | Insert |
| Baseline (ACT [11]) | 97±1.6 | 90±1.5 | 86±1.8 | 93±2.4 | 90±2.7 | 32±3.3 |
| EMA [4] + ACT | 98±0.8 | 96±1.8 | 96±1.8 | 92±1.5 | 90±1.5 | 40±2.3 |
| FE-EMA + ACT (ours) | 100±0.0 | 100±0.0 | **100**±0.0 | 94±1.5 | 94±1.5 | **58**±2.9 |
| TS-DWT + ACT (ours) | 100±0.0 | 98±1.3 | **98**±1.8 | 96±1.5 | 92±1.5 | **42**±2.3 |
| EMA + TS-DWT + ACT | 98±0.8 | 96±1.3 | 96±1.3 | 92±1.3 | 90±1.3 | 36±1.5 |
| FE-EMA + TS-DWT + ACT (ours: FEWT) | 100±0.0 | 98±0.8 | **98**±0.8 | 96±0.8 | 94±0.8 | **62**±1.5 |

To examine the interaction between the spatial and temporal-frequency modules, we further analyzed the marginal contribution of TS-DWT under different visual representation mechanisms. In the highly constrained Insert phase, adding TS-DWT to conventional EMA decreases the success rate from 40% to 36%, corresponding to a marginal change of −4 percentage points. In contrast, adding TS-DWT to FE-EMA increases the success rate from 58% to 62%, corresponding to a marginal improvement of +4 percentage points. The resulting 8-percentage-point contrast in the marginal effect of TS-DWT indicates that the frequency-aware visual representation generated by FE-EMA substantially improves its compatibility with subsequent temporal-frequency recalibration. In other words, FE-EMA changes the contribution of TS-DWT from detrimental under conventional spatial attention to beneficial under frequency-aligned spatial encoding.

This interaction is task dependent rather than uniformly additive. During the relatively smooth Lifted and Transfer phases, adding TS-DWT to FE-EMA changes the success rate from 100% to 98%, suggesting substantial representational redundancy when abrupt physical transitions are limited. In contrast, during the contact-rich Insert phase, the combined FE-EMA and TS-DWT configuration achieves the highest success rate of 62%, compared with 58% for FE-EMA alone, 42% for TS-DWT alone, and 32% for ACT. These results indicate that temporal-frequency recalibration becomes most beneficial when precise spatial alignment must be maintained while the policy simultaneously responds to abrupt contact-related changes. We therefore interpret the observed relationship as task-dependent spatial–temporal–frequency complementarity rather than a universally super-additive interaction. Each configuration reported in **Table 2** was independently repeated six times.

### 4.3. Grad-CAM Visualization

By utilizing Grad-CAM, we conduct a comparative visual analysis for tasks such as ***Cube Transfer*** and ***Bimanual Insertion***. As shown in **Fig. 7** and **Fig. 8**, after introducing the FE-EMA module into the ResNet18 backbone, the spatial high-response areas become significantly more concentrated on the task-relevant operational objects, avoiding attention dispersion to the background. This intuitive observation confirms that the FE-EMA module enhances the network's spatial filtering capabilities. Visualization of the dynamic weight $\alpha$ in the FE-EMA module see supplementary material **Fig. S2**.

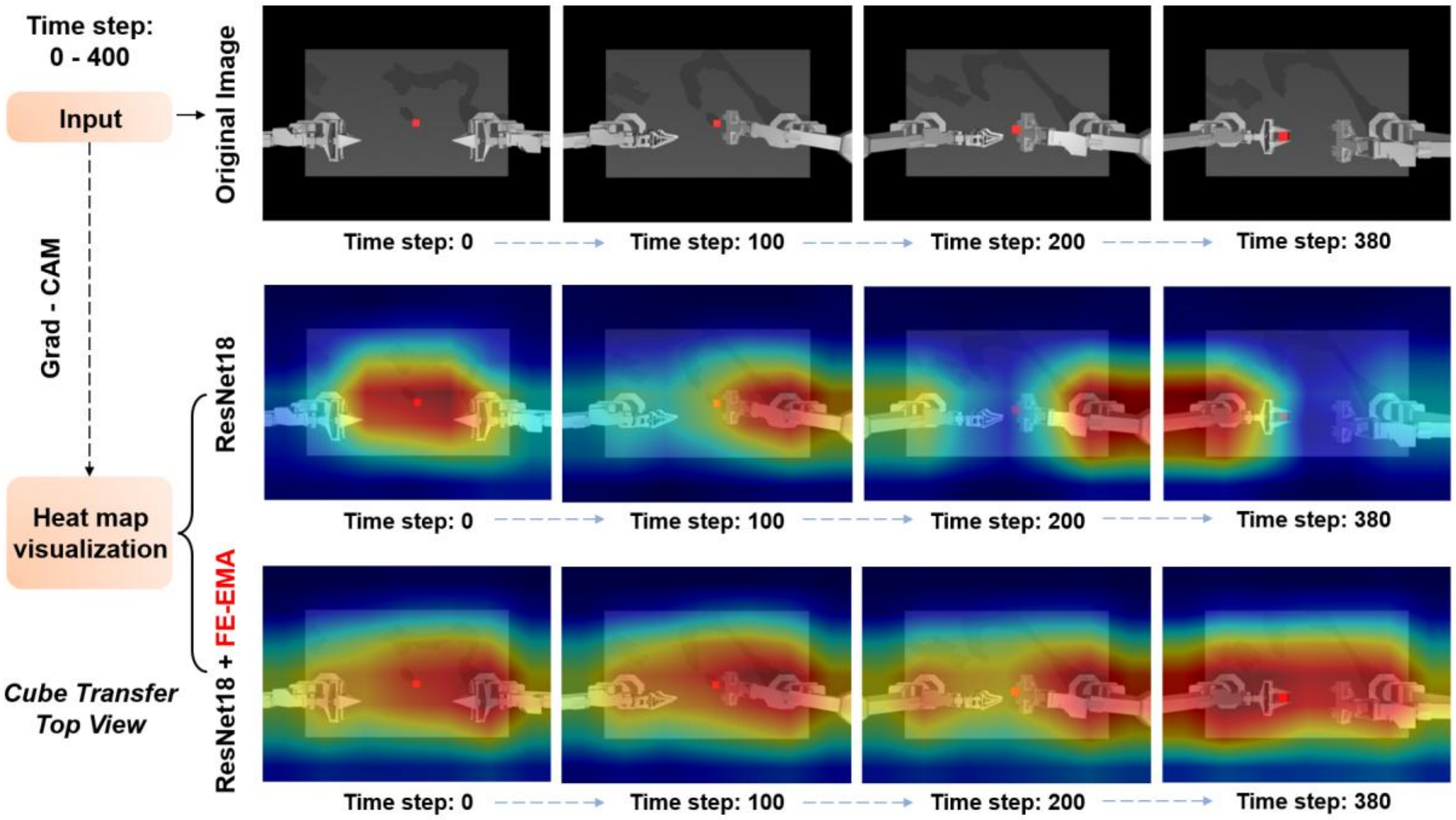

**Fig. 7.** Comparison of Heat map Visualizations for the ***Cube Transfer*** Task. Grad-CAM was utilized to visualize the model's attention regions, specifically for top view images.

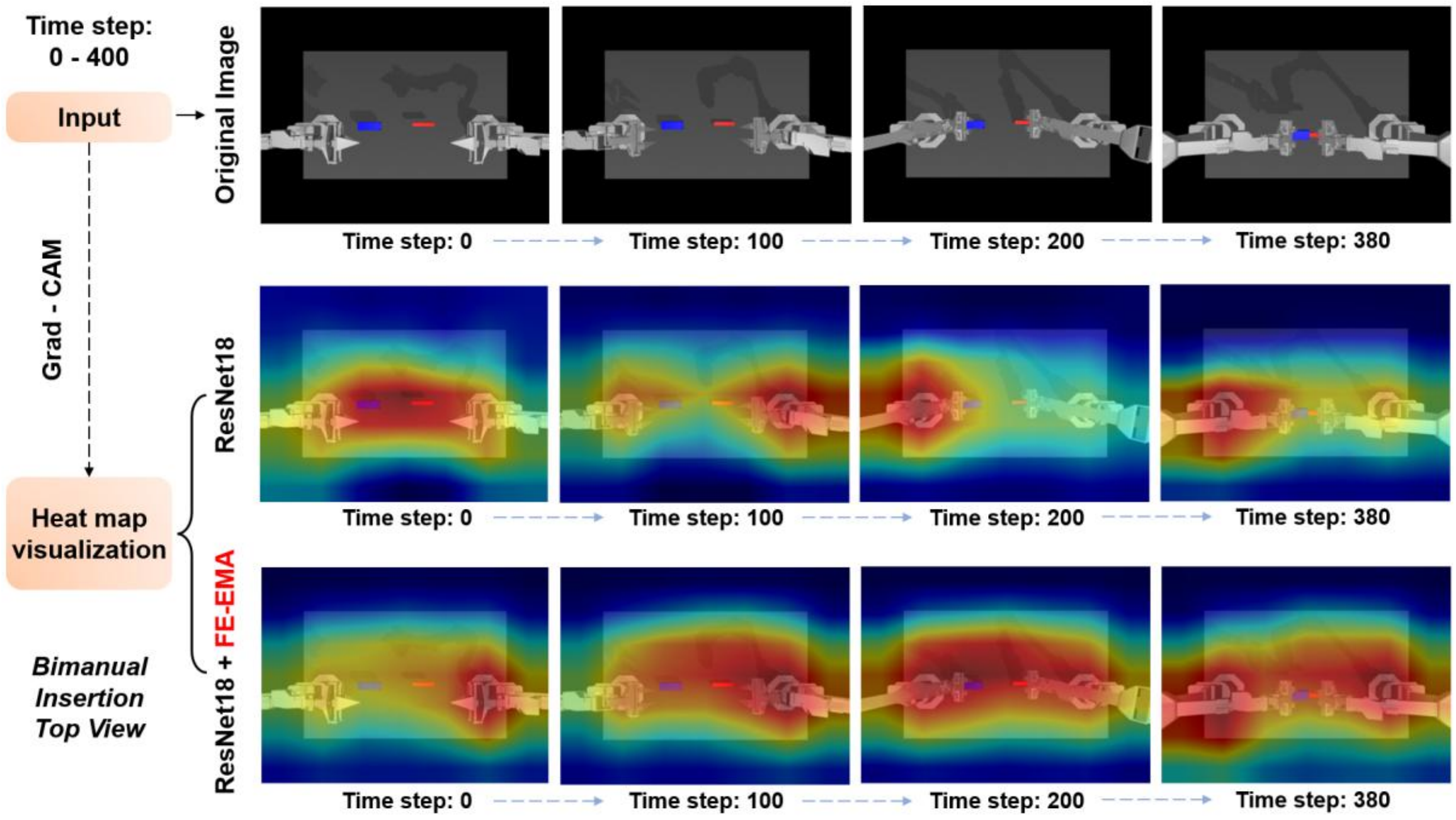

**Fig. 8.** Comparison of Heat map Visualizations for the ***Bimanual Insertion*** Task. Grad-CAM was utilized to visualize the model's attention regions, specifically for top view images.

In deep learning, interpretability analysis is highly beneficial for understanding model behavior. Grad-CAM is a widely used qualitative visualization method employed to reveal the key regions focused on by convolutional neural networks during feature extraction. It should be explicitly noted that in our study, Grad-CAM serves strictly as an auxiliary tool for qualitatively analyzing the attention mechanism of the FE-EMA module, entirely independent of the policy training process, and does not serve as a proof of correct action decision-making. This method utilizes the gradient information with respect to the last convolutional feature maps to perform a weighted fusion of channel responses, thereby generating a class activation heatmap. These heatmaps highlight the regions in the input image that contribute most significantly to the visual perception, effectively reflecting the model's attention mechanism. In short, the red regions in the Grad-CAM heatmaps simply indicate the areas where the model concentrates its attention during visual processing.

## 4.4. Robot Experiment

The superiority of the FEWT framework was verified through ablation experiments, followed by testing the skill learning of the robot using the FEWT model. **Table 3** summarizes the data of the robot in different simulated and real-world tasks. It compares the performance of the ACT and FEWT models in terms of task success. After the model training was completed, 50 tests were conducted for each task. The results show that FEWT (with IMU) can improve the ACT baseline by 6-12% in real-world mobile bimanual manipulation tasks. To highlight the advantages of the FEWT model in robot manipulation task success rates, this study selects mainstream robotic learning algorithms such as ACT and Diffusion Policy, and conducts comparative analysis of their performance in task completion success rates through both simulation and real-world experiments. Overall, the FEWT outperforms the existing mainstream methods, demonstrating superior performance. Detailed algorithm comparison results can be found in **Table 4**.

To ensure a rigorous and fair evaluation, all comparative experiments presented in Table 4 were strictly controlled. Specifically, our proposed FEWT and all baseline models (ACT and Diffusion Policy) were trained on the exact same expert demonstration datasets under comparable training budgets and identical computing resources. The hyperparameters for the baseline models were carefully tuned following the optimal configurations recommended in their respective original studies to ensure peak performance. Comprehensive details regarding the hyperparameter tuning, action horizons (chunk sizes), and network architectures for the Diffusion Policy baseline are thoroughly documented in **Table S4** of the Supplementary Material. **Table S5** summarizes the success rates at each task stage in simulation and real-world experiments. The chunk size sensitivity analysis and noise robustness test are shown in **Fig. S5** and **Fig. S6**, respectively.

**Table 3.** Summary of task-specific datasets and phases for robot manipulation in simulation and real-world

| Dataset (sim / real) | Phase1 | Phase2 | Phase3 | length | num | Total size | Success rate (ACT) | Success rate (FEWT) |
|---|---|---|---|---|---|---|---|---|
| *Building blocks* (sim) | grasp | place | place | 400 | 50 | 51.5G | 94±0.8 | 96±1.0 |
| *Drawer Storage* (sim) | grasp | lifted | place | 400 | 50 | 51.5G | 94±1.5 | 100±0.0 |
| *Lifted Curtain* (sim) | lifted | grasp | place | 350 | 50 | 45G | 84±1.5 | 92±1.3 |
| *Strip Transfer* (sim) | grasp | lifted | place | 300 | 50 | 38.6G | 96±0.8 | 100±0.0 |
| *Charm Splicing* (real) | lifted | approach | splicing | 800 | 20 | 41.2G | 92±1.5 | 98±1.5 |
| *Throw Paper* (real) | reach | grasp | place | 1000 | 25 | 64.4G | 86±1.5 | 94±1.5 |
| *Drawer Storage* (real) | reach | opened | place | 1200 | 25 | 77.2G | 74±1.5 | 84±1.5 |
| *Doll Storage* (real) | reach | transfer | place | 1200 | 25 | 77.2G | 70±1.5 | 82±2.3 |

**Table 4.** A comparative experiment on the success rates of different robot learning policy

| Policy | *Building blocks* (sim) | *Drawer Storage* (sim) | *Lifted Curtain* (sim) | *Strip Transfer* (sim) | *Charm Splicing* (real) | *Throw Paper* (real) | *Drawer Storage* (real) | *Doll Storage* (real) |
|---|---|---|---|---|---|---|---|---|
| DP (DDPM) [34] | 80±1.3 | 76±0.8 | 82±0.8 | 80±1.3 | 74±1.8 | 70±1.8 | 68±2.8 | 60±1.8 |
| DP (DDIM) [16] | 88±1.5 | 84±0.8 | 90±0.8 | 86±1.3 | 88±2.0 | 80±1.5 | 76±1.5 | 72±1.5 |
| ACT [11] (without IMU) | 94±0.8 | 94±1.5 | 84±1.5 | 96±0.8 | 92±1.5 | 86±1.5 | 74±1.5 | 70±1.5 |
| ACT [11] (with IMU) | – | – | – | – | 92±1.5 | 90±2.3 | 78±1.5 | 76±2.3 |
| FEWT (without IMU) | 96±1.0 | 100±0.0 | 92±1.3 | 100±0.0 | 98±1.5 | 90±1.8 | 80±2.0 | 76±1.5 |
| * FEWT (with IMU) | – | – | – | – | **98**±1.5 | **94**±1.5 | **84**±1.5 | **82**±2.3 |

* No chassis motion is involved in simulation; "–" indicates not evaluated; Each experiment was independently repeated six times.

Although the dataset size (20-25 real-world demonstrations per task) is relatively small compared to large-scale visual imitation learning, it is empirically sufficient for the FEWT framework. This high sample efficiency is fundamentally driven by the explicit frequency-domain enhancements (FE-EMA and TS-DWT). By isolating critical multi-scale features from redundant spatial background noise, the architecture prevents the model from over-relying on massive spatial datasets to infer physical dynamics. **Fig. 9** shows the static desktop skill learning screen of the robot in the MuJoCo simulation environment, along with the camera view for data collection of different tasks in a real-world scenario. In the MuJoCo simulation evaluations, the object initialization positions for each simulation task are randomly generated within a given interval.

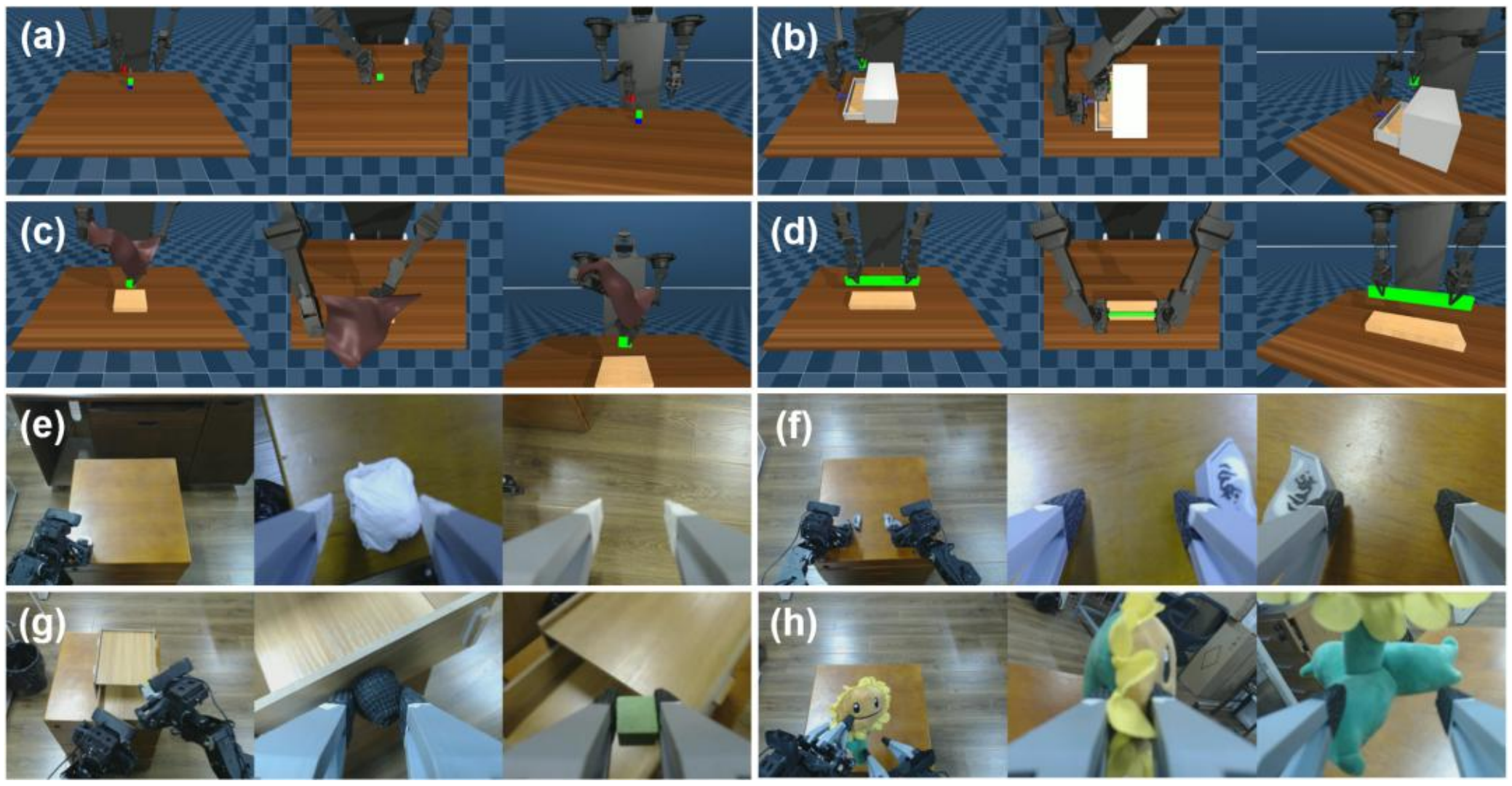

**Fig. 9.** Static desktop skill learning in the MuJoCo simulation environment with a camera view of data collection in the real-world. (a) *Building Blocks* (sim); (b) *Drawer Storage* (sim); (c) *Lifted Curtain* (sim); (d) *Strip Transfer* (sim); (e) *Throw Paper* (real); (f) *Charm Splicing* (real); (g) *Drawer Storage* (real); and (h) *Doll Storage* (real).

**Fig. 10** illustrates the effect of the robot autonomous learning manipulation in various real-world tasks, demonstrating the reliability and robustness of the FEWT model across these tasks. Within the evaluated mobile manipulation scenarios, FEWT enables coordinated whole-body control and demonstrates improved adaptability to platform-induced dynamic perturbations. In addition, the design of the robot is shown in supplementary material **Fig. S3**. Furthermore, to transparently analyze the current boundaries of the system, a detailed discussion and visualization of the representative failure cases (e.g., placement misalignments and grasping slips) encountered during these real-world autonomous evaluations are provided in supplementary material **Fig. S4**.

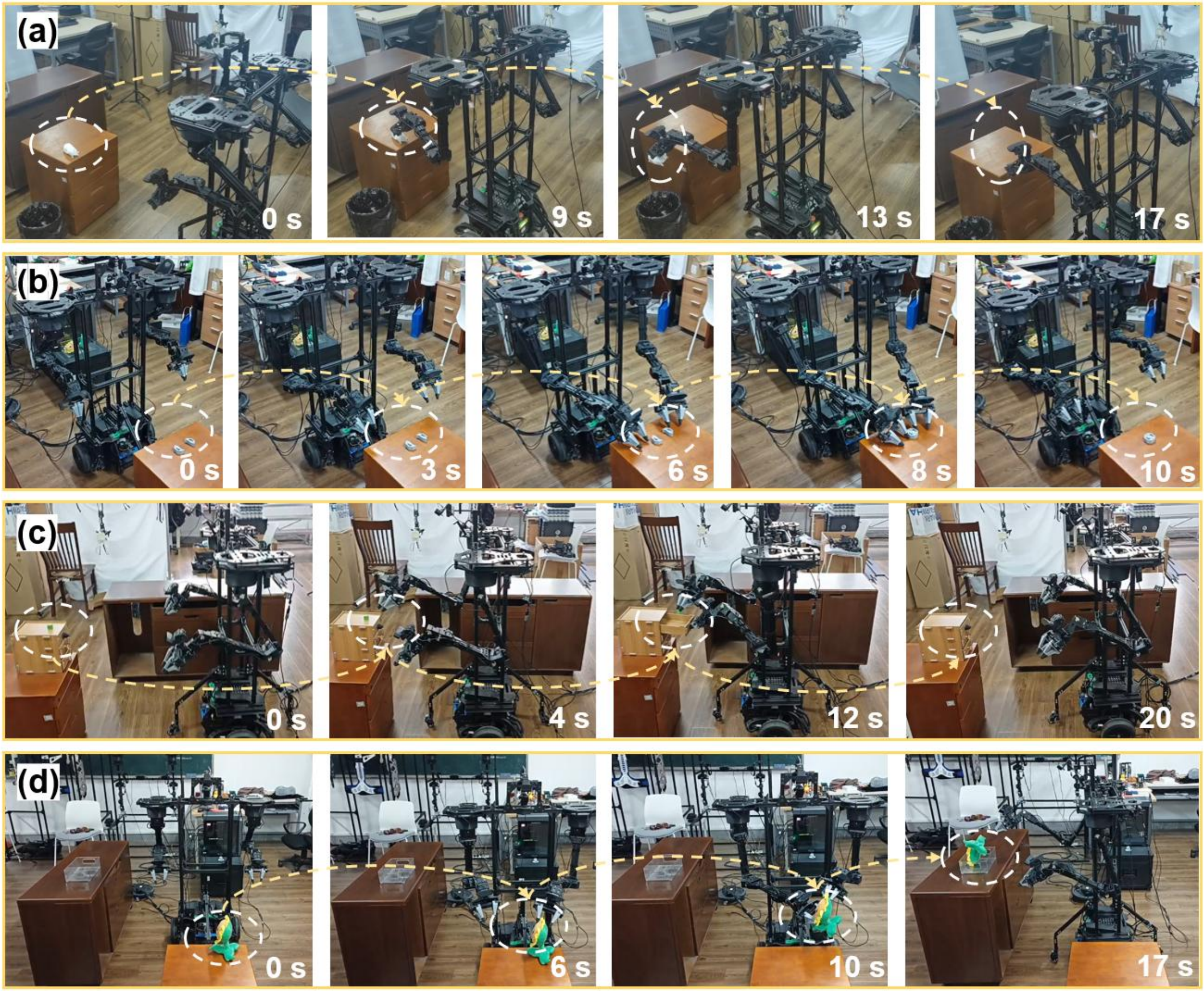

**Fig. 10.** Demonstration of skill learning by robot autonomous skills in the real-world. (a) *Throw Paper*; (b) *Charm Splicing*; (c) *Drawer Storage*; (d) *Doll Storage*.

Taking the tasks of *Building blocks* simulation and *Doll Storage* in real-world operations as examples, Grad-CAM is used for visual comparative analysis of the top images. **Fig. 11** presents a comparative analysis of the task (Top Image) visualization using Grad-CAM, applied to the simulation and real-world robot experiment. The model integrating the FE-EMA module significantly outperforms the model without the module in attention to the visual representation. The model integrated with the FE-EMA module shows a significant improvement in spatial focus in the Grad-CAM visualization results. Specifically, compared to the control model without this module, the high-response areas in its class activation maps are more concentrated on task-relevant operational goals or key structural regions, avoiding attention being dispersed to the background or irrelevant areas. This more discriminative spatial attention pattern indicates that the FE-EMA module effectively enhances the representational ability of convolutional features, indicating that the FE-EMA module effectively filters background spatial noise, allowing the convolutional backbone to maintain focus on task-relevant operational targets during feature extraction.

Furthermore, visual comparative analyses conducted across various simulation and real-robot tasks show consistent spatial focusing trends, corroborating the stable feature-extraction effectiveness of the module across different observation scenes.

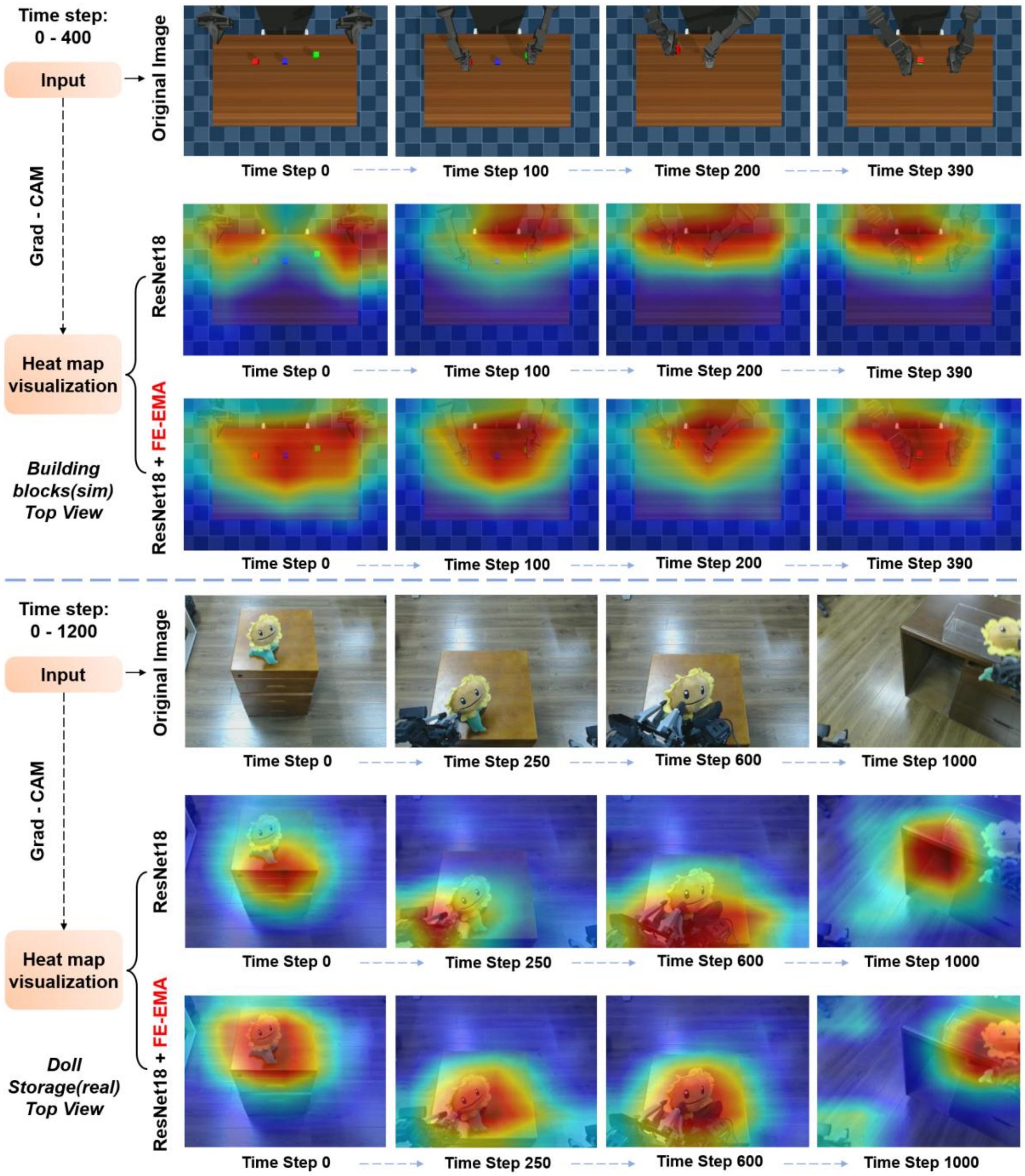

**Fig. 11.** Comparison of heatmap visualizations for the *Building blocks* simulation and *Doll Storage* in real-world task. Visualization of model regions of interest using Grad-CAM.

The Grad-CAM visualization technique qualitatively demonstrates the attention mechanism of the FE-EMA. For the qualitative analysis of the TS-DWT module, the attention of the frequency enhanced channels can be visualized. Compared to the output tensor of the Transformer encoder layer without TS-DWT integration, the FEWT framework is able to more significantly extract the importance of each channel dimension and the prominent features of different frequency components. The visualization of the frequency enhanced channel attention and the output tensor of the encoder layer on the Transformer are shown in **Fig. 12**.

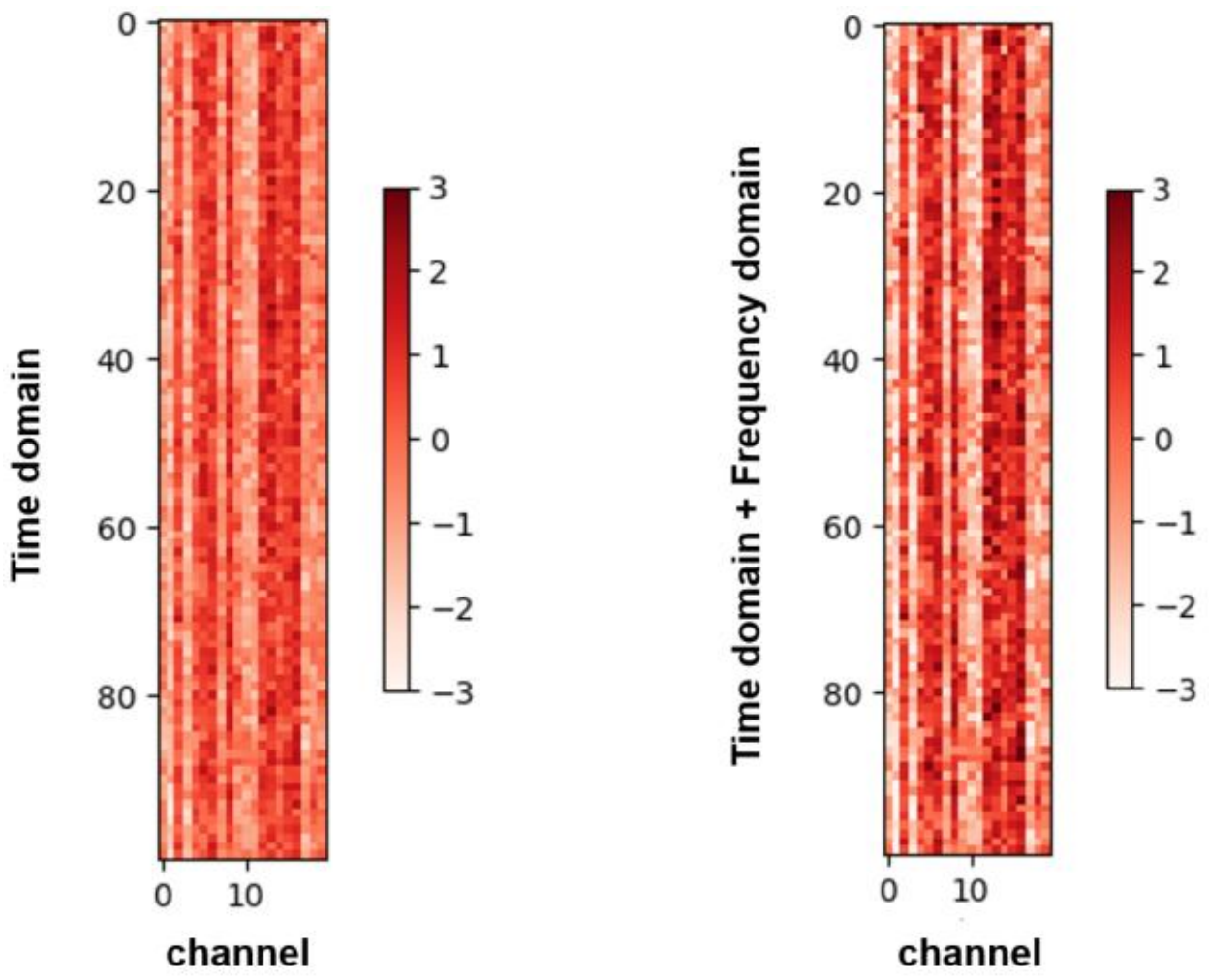

**Fig. 12.** Visualization of the frequency enhanced channel attention and the encoder layer's output tensor in the Transformer. $x$ axis represents the channel dimension and $y$ axis represents the frequency from low to high.

To systematically evaluate the performance differences among various algorithms in real-world robotic tasks, this study presents the success rate distributions based on 50 tests per algorithm in each experimental run. To enhance the statistical reliability and robustness of the conclusions, each set of experiments was independently repeated 6 times ($N$=6). Given the limited sample size per task, we adopted a highly transparent visualization protocol combining box plots overlaid with individual data strip plots. This approach comprehensively presents the raw data distribution, medians, and interquartile ranges, while directly exposing all underlying experimental variance. Furthermore, to provide sufficient quantitative support for assessing algorithmic stability, the non-parametric Mann-Whitney U test was employed to evaluate statistical significance. The resulting success rate comparisons, complete with significance brackets and $p$-values, are presented in **Fig. 13** (where the red values indicate the medians).

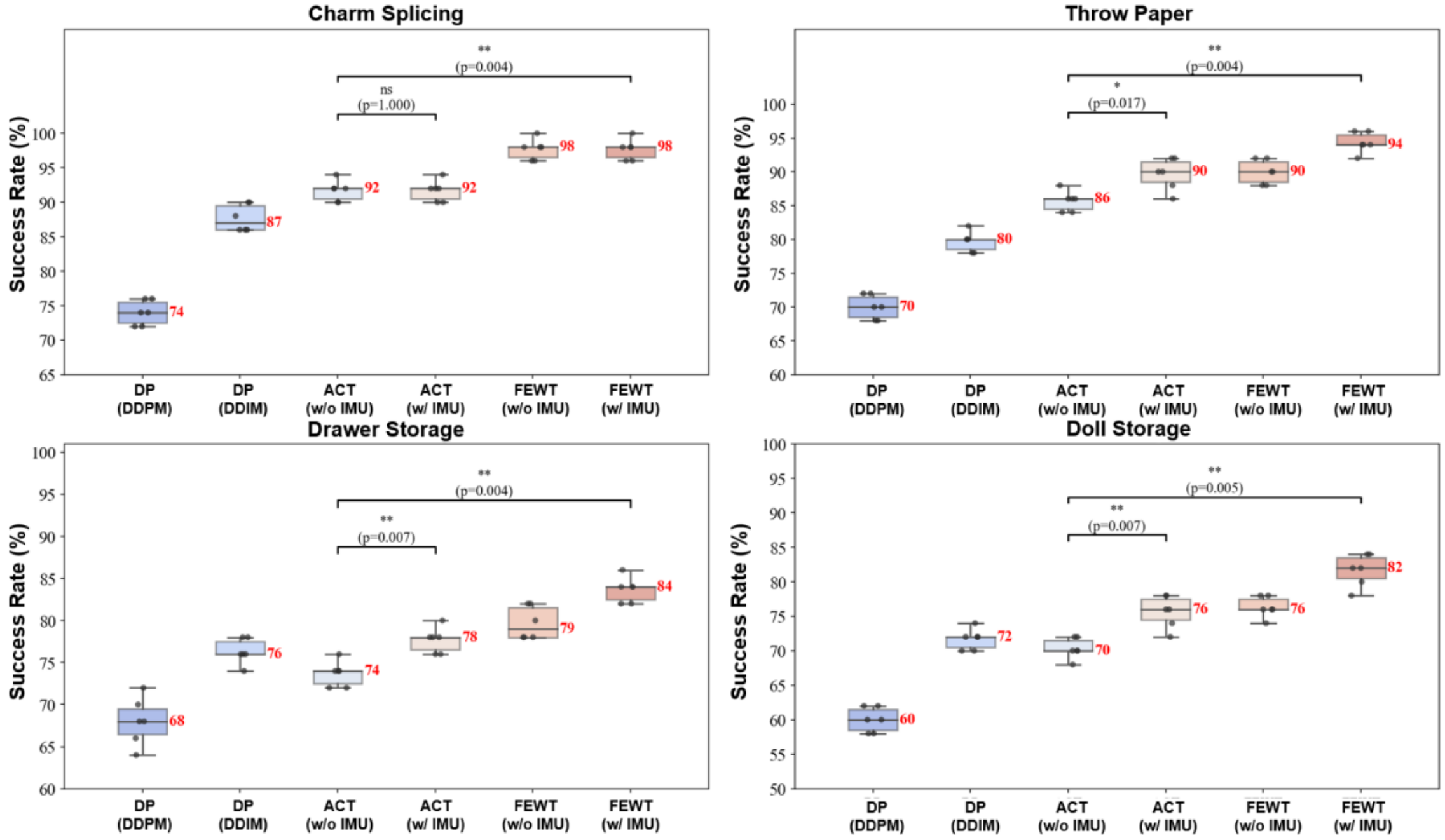

**Fig. 13.** Success rate comparison of algorithms in real-world robotic tasks box plots (ns: not significant, $*p < 0.05$, $**p < 0.01$).

### 4.5. Imitation learning with integrated tactile perception

Multimodal perception fusion is critical for enhancing imitation learning in contact-rich tasks. In this study, the introduction of the STF tactile sensor is not merely an addition of a hardware modality; fundamentally, it serves to provide a rich, high-frequency input source for the frequency-domain feature extraction of the FEWT algorithm. The STF-integrated grippers reconstruct the stress distribution on contact surfaces with high fidelity. Specifically, the 225-dimensional continuous tactile data acquired by the 15×15 array is fed directly into the network as high-frequency contact feature compensation. This multimodal fusion perfectly aligns with the spatial-temporal-frequency architecture of FEWT, significantly enhancing the model's adaptive correction capabilities against microscopic dynamic perturbations.

The detailed architecture for multimodal perception fusion and feature alignment is illustrated in **Fig. 14**. The pipeline is primarily divided into tactile signal preprocessing (left panel) and multimodal sequence inference (right panel). To ensure the robustness of tactile representations during complex physical interactions, the raw STF sensor data undergoes a rigorous signal conditioning pipeline: baseline background subtraction, temporal filtering for sequential noise suppression, Gaussian blur for spatial smoothing, and subsequent cropping with normalization. The processed high-fidelity 15×15 tactile tensor is then flattened into a 225-D feature vector and projected into a 512-D dense tactile embedding ($E_{tactile}$) via a Multi-Layer Perceptron (MLP). During the multimodal fusion phase, this high-frequency tactile embedding is concatenated with the vision embedding ($E_{vision}$, extracted via FE-EMA) and the proprioceptive embedding ($E_{prop}$). The fused feature vectors are combined with fixed position embeddings to formulate a comprehensive multimodal context sequence. This sequence is fed into the Transformer Encoder for global spatial-temporal modeling. The encoded memory is then processed by the TS-DWT, which performs multi-scale frequency decomposition and re-calibration along the feature dimension. Ultimately, the Transformer Decoder queries this frequency-enhanced multimodal memory to accurately predict the long-horizon robot action sequences.

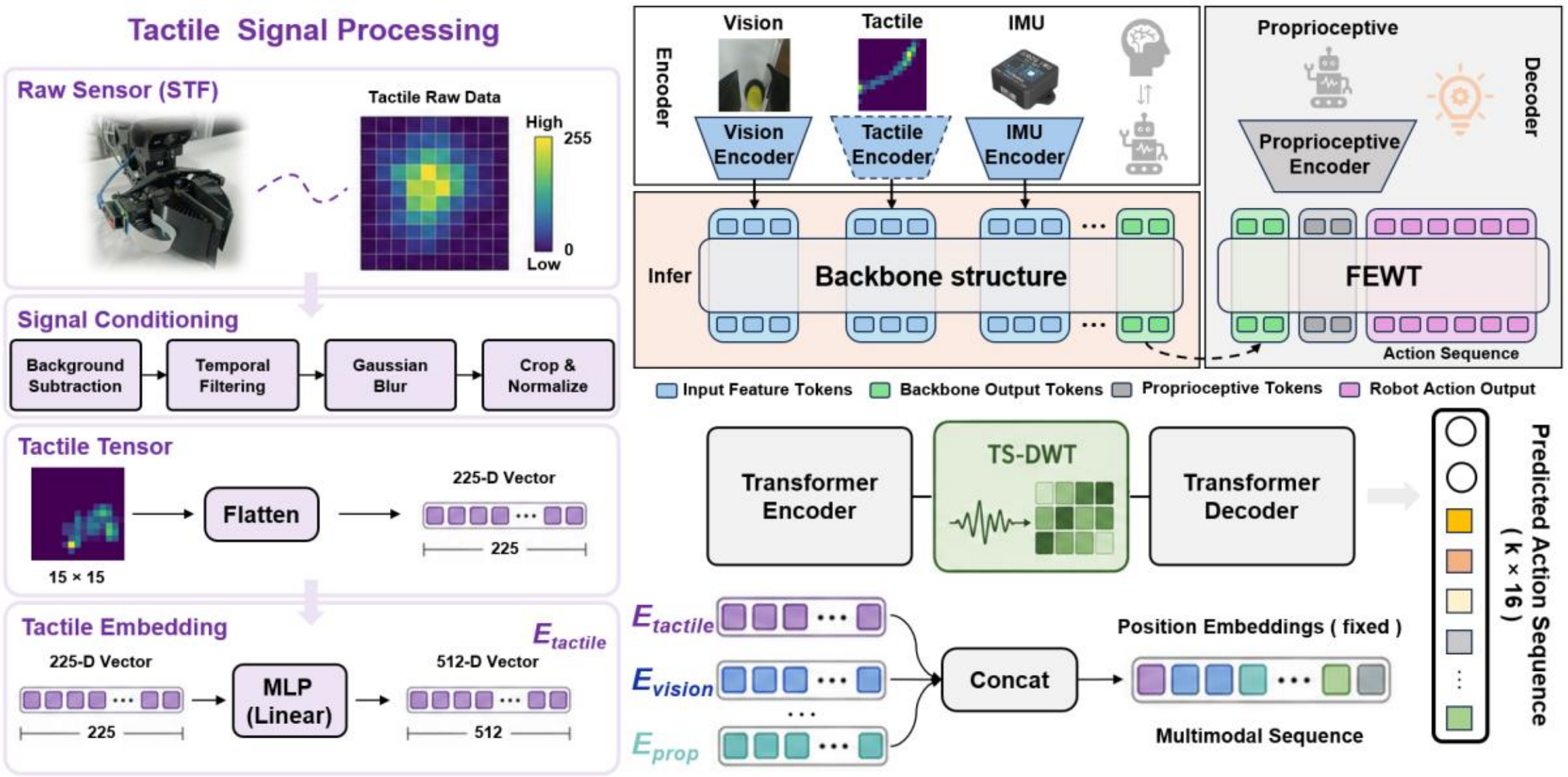

**Fig. 14.** Framework of the FEWT model incorporating tactile perception information.

The interaction among FE-EMA, TS-DWT, IMU feedback, and tactile sensing is hierarchical rather than a direct one-to-one coupling between individual sensors and network modules. FE-EMA first operates exclusively on the visual stream, preserving task-relevant spatial structures while introducing multiresolution frequency awareness into the visual tokens. In parallel, the joint states, chassis IMU signals, and processed tactile measurements are mapped through modality-specific projection layers into the same 512-dimensional latent space. These modality tokens are concatenated with the FE-EMA-enhanced visual tokens and jointly processed by the Transformer Encoder, allowing cross-modal dependencies to be learned before frequency recalibration. TS-DWT therefore does not process the raw IMU or tactile signals independently. Instead, it operates on the encoded multimodal memory in which visual, proprioceptive, inertial, and tactile information has already interacted through self-attention. Its function is to recalibrate the low- and high-frequency components of this fused representation before action decoding. Within this hierarchy, FE-EMA improves the spatial and frequency consistency of the visual stream, the IMU provides global platform-motion information, and STF supplies local end-effector contact information. TS-DWT subsequently reorganizes their fused representation across frequency sub-bands, enabling the Decoder to distinguish slowly varying task context from abrupt contact- and motion-related changes. The relative importance of these components consequently varies with the physical phase of the task. During free-space motion, visual and proprioceptive information largely determine the action trajectory. During mobile-base motion, inertial feedback becomes important for compensating for chassis acceleration, angular motion, vibration, and drift. During grasping, transfer, and insertion, tactile stress distributions become increasingly informative because contact stability, slip, and collision cannot be uniquely inferred from visual proximity alone. Thus, the improvement of FEWT arises from complementary information being introduced at the appropriate physical scale and subsequently integrated within a common frequency-aware representation.

During real-world deployment, the FEWT model with tactile perception was evaluated on the wheeled bimanual robot platform. For the long-horizon mobile manipulation tasks of *Object Storage* and *Bimanual Carry* (50Hz control frequency, 1500 control steps), **Fig. 15** illustrates the strict temporal alignment between the global vision and bimanual tactile features. To ensure intuitive and synchronous observation, the real-time tactile stress distributions of both the left and right arms are explicitly distinguished yet merged within a single, unified visualization screen.

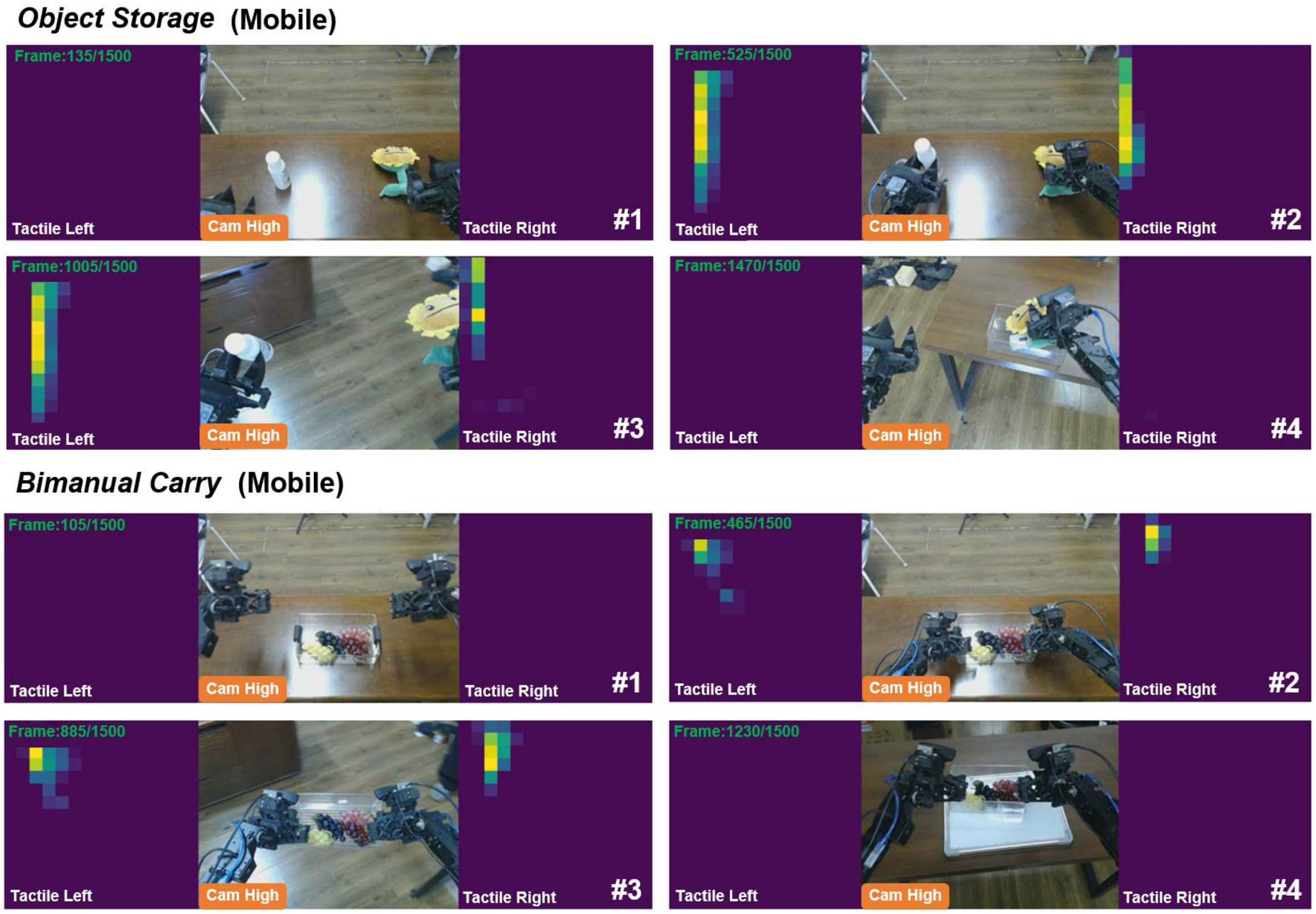

**Fig. 15.** Two typical mobile manipulation tasks of *Object Storage* and *Bimanual Carry*.

To further highlight the critical role of tactile feedback in dexterous manipulation, we conducted *Bimanual Transfer* and *Bimanual Insertion* tasks (**Fig. 16**). As shown in the quantitative results of **Table 5**, benefiting from the high-resolution, high-frequency stress compensation provided by the STF, the FEWT model demonstrated a commanding advantage in assembly tasks requiring extreme micro-manipulation precision, achieving success rates significantly superior to existing baseline models. These experiments thoroughly validate that the FEWT model, empowered by high-frequency tactile compensation, ensures the robust and precise execution of complex, long-horizon tasks.

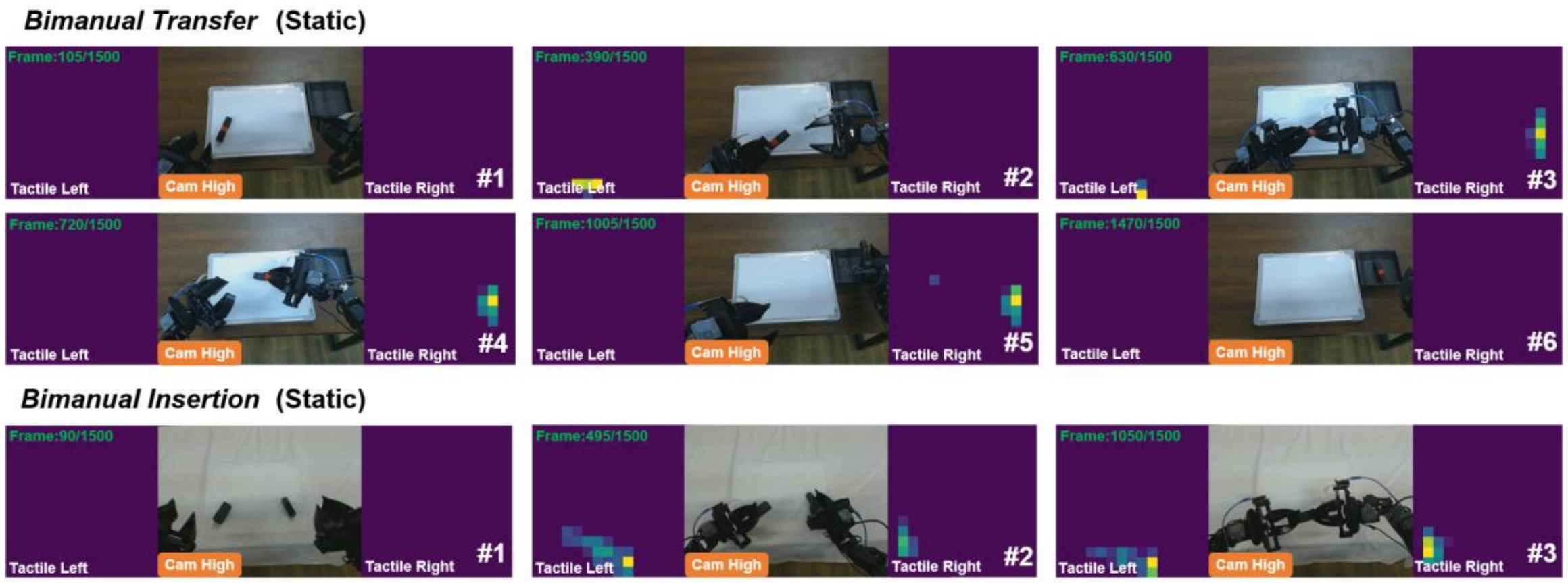

**Fig. 16.** Deployment performance of the FEWT model in *Bimanual Transfer* and *Bimanual Insertion* tasks.

The quantitative results in **Table 5** show that, after incorporating tactile perception, the proposed FEWT model achieves a significant improvement in task success rate compared with multiple baseline models. This advantage is mainly attributed to the integration of tactile sensory data. The STF provides high-resolution stress perception, enabling the model to more accurately capture contact states and local force variations, thereby demonstrating clear advantages in *Bimanual Insertion* tasks that require high-precision manipulation. The ablation experiment on the effect of IMU integration on the real-world performance of mobile whole-body control manipulation is presented in **Table S6**. Each experiment was independently repeated six times.

**Table 5.** Comparison of real-world experimental success rates for robots integrated with tactile perception

| Method (with IMU) | *Object Storage* (Mobile) | *Bimanual Carry* (Mobile) | *Bimanual Transfer* (Static) | *Bimanual Insertion* (Static) | Overall Success Rate |
|---|---|---|---|---|---|
| DDIM [16] | 52±1.5 | 54±1.5 | 72±1.6 | 24±1.3 | 50.8±1.3 |
| DDIM +STF | 56±1.3 | 58±1.0 | 78±0.8 | 30±1.5 | 55.8±0.9 |
| ACT [11] | 50±1.5 | 56±1.5 | 74±2.2 | 28±1.8 | 52.2±1.7 |
| ACT+STF | 58±1.0 | 62±1.5 | 82±1.5 | 34±1.5 | 59.4±1.3 |
| FEWT (ours) | **94**±1.0 | **92**±1.5 | **96**±1.5 | **76**±1.0 | **90.0**±1.2 |
| FEWT+STF (ours) | **98**±0.8 | **98**±1.0 | **100**±0.0 | **82**±1.8 | **94.8**±0.8 |

Because IMU feedback was retained for every configuration in **Table 5**, these comparisons quantify the conditional contribution of tactile sensing after global platform-motion information is already available. For FEWT, adding STF improves *Object Storage*, *Bimanual Carry*, *Bimanual Transfer*, and *Bimanual Insertion* by 4, 6, 4, and 6 percentage points, respectively, and increases the overall success rate from 90.0% to 94.8%.

The improvement is observed in both mobile and static tasks, but its physical interpretation differs across task types. In *Bimanual Carry*, tactile feedback helps maintain stable bimanual contact while the mobile base introduces motion-induced variations. In *Bimanual Insertion*, it provides local contact-state information required to distinguish successful alignment, collision, and unstable contact. These results indicate that STF complements, rather than duplicates, IMU feedback: IMU describes global platform dynamics, whereas STF directly observes local end-effector interactions.

The experimental results demonstrate that, by integrating tactile sensors, the system can leverage high-fidelity contact-surface stress perception to adaptively correct microscopic dynamic variations during manipulation, thereby ensuring compliant and stable end-effector motions in bimanual operations. The introduction of tactile perception effectively enhances the robot's ability to perceive contact states and local force variations, further improving the robustness and precision of complex long-horizon task execution.

The real-world experiments also provide evidence of robustness under several practically relevant challenging conditions within the learned task distribution. In the mobile *Object Storage* and *Bimanual Carry* tasks, the policy must coordinate bimanual manipulation with wheeled-base motion, thereby continuously encountering chassis acceleration, vibration, wheel slip, odometry drift, and viewpoint variation. As further quantified by the IMU ablation study in Supplementary **Table S6**, incorporating IMU feedback improves the overall success rate of FEWT from 88.2% to 98.3%, indicating that the multimodal policy can effectively compensate for the dynamic perturbations induced by whole-body mobile manipulation. In addition, the *Drawer Storage* and *Doll Storage* tasks (**Fig. 9**) naturally involve intermittent visual occlusions caused by the manipulated objects and mutual blockage between the two robotic arms. These observations provide qualitative evidence that FEWT can tolerate moderate and transient occlusions when complementary visual, proprioceptive, and tactile information remains available. However, severe or persistent occlusion may remove critical geometric cues from all camera views and still lead to failure, as transparently analyzed in Supplementary **Fig. S4**. Therefore, the robustness demonstrated here should be understood within the evaluated task and disturbance distributions, rather than as invariance to arbitrary open-world visual changes. Additionally, cross-component interactions are summarized in Supplementary **Table S7**.

## 5. Conclusions

This study presented a custom-built wheeled bimanual robotic platform equipped with an exoskeleton-style teleoperation system, which effectively facilitated the collection of high-quality, anthropomorphic expert demonstrations. To address the inherent limitations of pure spatial representations in complex dynamic tasks, we proposed a novel imitation learning framework termed the Frequency-Enhanced Wavelet-based Transformer (FEWT). The core innovation of FEWT lies in its holistic spatial-temporal-frequency architecture. By synergizing the Frequency-Enhanced Efficient Multi-Scale Attention (FE-EMA) and Time-Series Discrete Wavelet Transform (TS-DWT) modules, the framework explicitly extracts and aligns multi-scale features, effectively filtering background noise while capturing critical high-frequency dynamics. Furthermore, the seamless integration of a customized flexible sensor, the Smart Tactile Fabric (STF), provided high-fidelity, high-frequency stress compensation. This multimodal fusion significantly enhanced the system's adaptive robustness against microscopic kinematic perturbations during contact-rich operations.

Experimental evaluations demonstrated the superiority of our approach. FEWT significantly outperformed state-of-the-art baselines, improving task success rates by up to 30 percentage points in the most challenging simulated insertion scenarios, and by 6 to 12 percentage points across real-world evaluations. Future work will integrate web-scale Vision-Language-Action representations to improve unseen-object generalization, photometric and geometric augmentation to address illumination and viewpoint shifts, active multi-view or 3D perception to mitigate severe occlusion, and recovery-from-error demonstrations and hierarchical policy structures to support extended autonomous operation in dynamic environments.

## 6. Limitations

A limitation of the current task-specific imitation learning setup is its reliance on the trained object categories. Generalization to entirely novel, unseen semantic objects remains challenging without leveraging web-scale pre-trained models, which constitutes a primary direction for our future research. Furthermore, the real-world tasks evaluated in this study serve primarily as proof-of-concept demonstrations and do not yet fully capture the immense complexity of unconstrained industrial assembly or dynamic household environments. Evaluating the system's robustness under severe external disturbances and scaling to more complex, long-horizon tasks—ultimately achieving several hours or even a full day of automated, closed-loop operation without human intervention—represent critical next steps for our future empirical validations.

Finally, due to the strict hardware-algorithm coupling in end-to-end imitation learning, directly evaluating the proposed method on publicly available manipulation benchmarks is currently infeasible. Although the core FEWT modules are not mathematically tied to a specific robotic embodiment, the present experiments were conducted exclusively on our custom-built wheeled bimanual platform. Therefore, cross-platform transferability has not yet been empirically established. FE-EMA operates on generic visual feature maps, and TS-DWT processes Transformer memory representations independently of the robot kinematic structure. The proprioceptive, IMU, tactile, and action spaces are connected to the core policy through modality-specific input projections and an embodiment-dependent output head. In principle, adapting FEWT to another platform would therefore require reconfiguring these interface layers and retraining or fine-tuning the policy, rather than redesigning the core architecture. However, differences in morphology, sensor placement, action dimensionality, control frequency, and data distribution may still produce substantial embodiment gaps. Systematic evaluation on additional robotic platforms and compatible public manipulation datasets will be an important direction of future work.

The robustness claims of the present study are limited to the evaluated conditions, including randomized object initialization within predefined task ranges, mobile-base-induced kinematic disturbances, injected spatial perturbations, and multi-stage manipulation sequences. They should not be interpreted as evidence of zero-shot generalization to entirely unseen semantic objects or invariance to arbitrary illumination changes, persistent visual occlusions, and unconstrained environments containing independently moving external agents. The current FEWT framework remains a task-specific imitation-learning policy, and these broader out-of-distribution settings would require substantially more diverse demonstrations, large-scale pre-trained representations, explicit domain randomization, and recovery-oriented policy learning.

## Conflicts of Interest

The authors declare no competing interests.

## References

[1] M. Zare, P. M. Kebria, A. Khosravi, and S. Nahavandi, "A survey of imitation learning: Algorithms, recent developments, and challenges," *IEEE Transactions on Cybernetics,* vol. 54, no. 12, pp. 7173-7186, 2024. https://doi.org/10.1109/tcyb.2024.3395626.

[2] X. Chen, S. Xie, and K. He, "An empirical study of training self-supervised vision transformers," in *Proceedings of the IEEE/CVF international conference on computer vision*, 2021, pp. 9640-9649. https://doi.org/10.1109/iccv48922.2021.00950.

[3] K. Burns, Z. Witzel, J. I. Hamid, T. Yu, C. Finn, and K. Hausman, "What Makes Pre-Trained Visual Representations Successful for Robust Manipulation?," in *8th Annual Conference on Robot Learning*, 2024.

[4] D. Ouyang, S. He, G. Zhang, *et al.*, "Efficient multi-scale attention module with cross-spatial learning," in *ICASSP 2023-2023 IEEE international conference on acoustics, speech and signal processing (ICASSP)*, 2023: IEEE, pp. 1-

5. https://doi.org/10.1109/icassp49357.2023.10096516.
[5] K. He, X. Zhang, S. Ren, and J. Sun, “Deep residual learning for image recognition,” in *Proceedings of the IEEE conference on computer vision and pattern recognition*, 2016, pp. 770-778. https://doi.org/10.1109/cvpr.2016.90.
[6] G. Xu, W. Liao, X. Zhang, C. Li, X. He, and X. Wu, “Haar wavelet downsampling: A simple but effective downsampling module for semantic segmentation,” *Pattern recognition,* vol. 143, p. 109819, 2023. https://doi.org/10.1016/j.patcog.2023.109819.
[7] Y. Yang, G. Yuan, and J. Li, “SFFNet: A wavelet-based spatial and frequency domain fusion network for remote sensing segmentation,” *IEEE Transactions on Geoscience and Remote Sensing,* vol. 62, pp. 1-17, 2024. https://doi.org/10.1109/tgrs.2024.3427370.
[8] C. Zhao, W. Cai, C. Dong, and C. Hu, “Wavelet-based fourier information interaction with frequency diffusion adjustment for underwater image restoration,” in *Proceedings of the IEEE/CVF conference on computer vision and pattern recognition*, 2024, pp. 8281-8291. https://doi.org/10.1109/cvpr52733.2024.00791.
[9] M. Jiang, P. Zeng, K. Wang, H. Liu, W. Chen, and H. Liu, “FECAM: Frequency enhanced channel attention mechanism for time series forecasting,” *Advanced Engineering Informatics,* vol. 58, p. 102158, 2023. https://doi.org/10.1016/j.aei.2023.102158.
[10] R. R. Selvaraju, M. Cogswell, A. Das, R. Vedantam, D. Parikh, and D. Batra, “Grad-cam: Visual explanations from deep networks via gradient-based localization,” in *Proceedings of the IEEE international conference on computer vision*, 2017, pp. 618-626. https://doi.org/10.1109/iccv.2017.74.
[11] T. Zhao, V. Kumar, S. Levine, and C. Finn, “Learning Fine-Grained Bimanual Manipulation with Low-Cost Hardware,” *Robotics: Science and Systems XIX,* 2023. https://doi.org/10.15607/rss.2023.xix.016.
[12] E. Todorov, T. Erez, and Y. Tassa, “Mujoco: A physics engine for model-based control,” in *2012 IEEE/RSJ international conference on intelligent robots and systems*, 2012: IEEE, pp. 5026-5033. https://doi.org/10.1109/iros.2012.6386109.
[13] A. Vaswani, N. Shazeer, N. Parmar, *et al.*, “Attention is all you need,” *Advances in neural information processing systems,* vol. 30, 2017.
[14] K. Darvish *et al.*, “Teleoperation of humanoid robots: A survey,” *IEEE Transactions on Robotics,* vol. 39, no. 3, pp. 1706-1727, 2023. https://doi.org/10.1109/tro.2023.3236952.
[15] A. Purushottam, Y. Jung, K. Murphy, D. Baek, and J. Ramos, “Hands-free telelocomotion of a wheeled humanoid,” in *2022 IEEE/RSJ International Conference on Intelligent Robots and Systems (IROS)*, 2022: IEEE, pp. 8313-8320. https://doi.org/10.1109/iros47612.2022.9981629.
[16] Z. Fu, T. Z. Zhao, and C. Finn, “Mobile aloha: Learning bimanual mobile manipulation using low-cost whole-body teleoperation,” in *8th Annual Conference on Robot Learning*, 2024.
[17] Y. Liang, W. Li, Y. Wang, R. Xiong, Y. Mao, and J. Zhang, “Dynamic movement primitive based motion retargeting for dual-arm sign language motions,” in *2021 IEEE International Conference on Robotics and Automation (ICRA)*, 2021: IEEE, pp. 8195-8201. https://doi.org/10.1109/icra48506.2021.9561120.
[18] K. Ayusawa and E. Yoshida, “Motion retargeting for humanoid robots based on simultaneous morphing parameter identification and motion optimization,” *IEEE Transactions on Robotics,* vol. 33, no. 6, pp. 1343-1357, 2017. https://doi.org/10.1109/tro.2017.2752711.
[19] C. Chi, Z. Xu, C. Pan, *et al.*, “Universal Manipulation Interface: In-The-Wild Robot Teaching Without In-The-Wild Robots,” *Robotics: Science and Systems XX,* 2024. https://doi.org/10.15607/rss.2024.xx.045.
[20] C. Zhou, L. Zhao, H. Wang, L. Chen, and Y. Zheng, “A bilateral dual-arm teleoperation robot system with a unified control architecture,” in *2021 30th IEEE International Conference on Robot & Human Interactive Communication (RO-MAN)*, 2021: IEEE, pp. 495-502. https://doi.org/10.1109/ro-man50785.2021.9515523.
[21] M. Schwarz, C. Lenz, A. Rochow, M. Schreiber, and S. Behnke, “Nimbro avatar: Interactive immersive telepresence with force-feedback telemanipulation,” in *2021 IEEE/RSJ International Conference on Intelligent Robots and Systems (IROS)*, 2021: IEEE, pp. 5312-5319. https://doi.org/10.1109/iros51168.2021.9636191.
[22] A. Iyer, Z. Peng, Y. Dai, *et al.*, “Open teach: A versatile teleoperation system for robotic manipulation,” *arXiv preprint arXiv:2403.07870,* 2024.
[23] A. Toedtheide, X. Chen, H. Sadeghian, A. Naceri, and S. Haddadin, “A force-sensitive exoskeleton for teleoperation: An application in elderly care robotics,” in *2023 IEEE International Conference on Robotics and Automation (ICRA)*, 2023: IEEE, pp. 12624-12630. https://doi.org/10.1109/icra48891.2023.10161175.
[24] Y. Ishiguro *et al.*, “Bilateral humanoid teleoperation system using whole-body exoskeleton cockpit tablis,” *IEEE Robotics and Automation Letters,* vol. 5, no. 4, pp. 6419-6426, 2020. https://doi.org/10.1109/lra.2020.3013863.
[25] H. Fang, HS. Fang, Y. Wang, *et al.*, “Airexo: Low-cost exoskeletons for learning whole-arm manipulation in the wild,” in *2024 IEEE International Conference on Robotics and Automation (ICRA)*, 2024: IEEE, pp. 15031-15038. https://doi.org/10.1109/icra57147.2024.10610799.
[26] S. Yang, M. Liu, Y. Qin, *et al.*, “ACE: A Cross-platform and visual-Exoskeletons System for Low-Cost Dexterous Teleoperation,” in *8th Annual Conference on Robot Learning*, 2024.
[27] A. Gupta, S. Savarese, S. Ganguli, and L. Fei-Fei, “Embodied intelligence via learning and evolution,” *Nature communications,* vol. 12, no. 1, p. 5721, 2021. https://doi.org/10.1038/s41467-021-25874-z.
[28] H. Kim, Y. Ohmura, and Y. Kuniyoshi, “Goal-conditioned dual-action imitation learning for dexterous dual-arm robot manipulation,” *IEEE Transactions on Robotics,* vol. 40, pp. 2287-2305, 2024.

https://doi.org/10.1109/tro.2024.3372778.
[29] J. Gao, X. Jin, F. Krebs, N. Jaquier, and T. Asfour, "Bi-kvil: Keypoints-based visual imitation learning of bimanual manipulation tasks," in *2024 IEEE international conference on robotics and automation (ICRA)*, 2024: IEEE, pp. 16850-16857. https://doi.org/10.1109/icra57147.2024.10610763.
[30] D. A. Pomerleau, "Alvinn: An autonomous land vehicle in a neural network," *Advances in neural information processing systems,* vol. 1, 1988.
[31] A. Brohan, N. Brown, J. Carbajal, *et al.*, "Rt-1: Robotics transformer for real-world control at scale," *arXiv preprint arXiv:2212.06817,* 2022.
[32] B. Zitkovich, T. Yu, S. Xu, *et al.*, "Rt-2: Vision-language-action models transfer web knowledge to robotic control," *arXiv preprint arXiv:2307.15818*, 2023.
[33] A. O'Neill, A. Rehman, A. Maddukuri, *et al.*, "Open x-embodiment: Robotic learning datasets and rt-x models: Open x-embodiment collaboration$^0$," in *2024 IEEE International Conference on Robotics and Automation (ICRA)*, 2024: IEEE, pp. 6892-6903. https://doi.org/10.1109/icra57147.2024.10611477.
[34] C. Chi *et al.*, "Diffusion policy: Visuomotor policy learning via action diffusion," *The International Journal of Robotics Research,* vol. 44, no. 10-11, pp. 1684-1704, 2025. https://doi.org/10.1177/02783649241273668.
[35] J. Ho, A. Jain, and P. Abbeel, "Denoising diffusion probabilistic models," *Advances in neural information processing systems,* vol. 33, pp. 6840-6851, 2020.
[36] Y. Ze, G. Zhang, K. Zhang, C. Hu, M. Wang, and H. Xu, "3D Diffusion Policy: Generalizable Visuomotor Policy Learning via Simple 3D Representations," *Robotics: Science and Systems XX,* 2024. https://doi.org/10.15607/rss.2024.xx.067.
[37] Q. Lv, H. Li, X. Deng, *et al.*, "Spatial-temporal graph diffusion policy with kinematic modeling for bimanual robotic manipulation," in *Proceedings of the Computer Vision and Pattern Recognition Conference*, 2025, pp. 17394-17404. https://doi.org/10.1109/cvpr52734.2025.01621.
[38] L. X. Shi, A. Sharma, T. Z. Zhao, and C. Finn, "Waypoint-Based Imitation Learning for Robotic Manipulation," in *7th Annual Conference on Robot Learning*, 2023.
[39] Z. Xu *et al.*, "UniT: Data Efficient Tactile Representation With Generalization to Unseen Objects," *IEEE Robotics and Automation Letters,* vol. 10, no. 6, pp. 5481-5488, 2025. https://doi.org/10.1109/lra.2025.3559835.
[40] B. Huang, Y. Wang, X. Yang, Y. Luo, and Y. Li, "3D-ViTac: Learning Fine-Grained Manipulation with Visuo-Tactile Sensing," in *8th Annual Conference on Robot Learning*, 2024.
[41] B. Huang, J. Xu, I. Akinola, *et al.*, "VT-Refine: Learning Bimanual Assembly with Visuo-Tactile Feedback via Simulation Fine-Tuning," in *9th Annual Conference on Robot Learning*, 2025.
[42] X. Zhu, B. Huang, and Y. Li, "Touch in the Wild: Learning Fine-Grained Manipulation with a Portable Visuo-Tactile Gripper," In *Advances in Neural Information Processing Systems (NeurIPS)* 38, 2025, pp. 170711–170740. Neural Information Processing Systems Foundation, Inc. https://doi.org/10.52202/085713-5143.
[43] N. Gu, K. Kosuge, and M. Hayashibe, "TactileAloha: Learning Bimanual Manipulation With Tactile Sensing," *IEEE Robotics and Automation Letters,* vol. 10, no. 8, pp. 8348-8355, 2025. https://doi.org/10.1109/lra.2025.3585396.
[44] S. Kareer, D. Patel, R. Punamiya, *et al.*, "EgoMimic: Scaling Imitation Learning via Egocentric Video," in *2025 IEEE International Conference on Robotics and Automation (ICRA)*, 19-23 May 2025 2025, pp. 13226-13233. https://doi.org/10.1109/icra55743.2025.11127989.
[45] L. Y. Zhu *et al.*, "EMMA: Scaling Mobile Manipulation via Egocentric Human Data," *IEEE Robotics and Automation Letters,* vol. 11, no. 3, pp. 3087-3094, 2026. https://doi.org/10.1109/lra.2026.3653320.
[46] Y. Wang *et al.*, "Vla-adapter: An effective paradigm for tiny-scale vision-language-action model," in *Proceedings of the AAAI conference on artificial intelligence*, 2026, vol. 40, no. 22, pp. 18638-18646. https://doi.org/10.1609/aaai.v40i22.38931.
[47] M. J. Kim, K. Pertsch, S. Karamcheti, *et al.*, "OpenVLA: An Open-Source Vision-Language-Action Model," in *8th Annual Conference on Robot Learning*, 2024.
[48] M. J. Kim, C. Finn, and P. Liang, "Fine-tuning vision-language-action models: Optimizing speed and success," *arXiv preprint arXiv:2502.19645,* 2025.
[49] K. Black, N. Brown, D. Driess, *et al.*, "$\pi_0$: A Vision-Language-Action Flow Model for General Robot Control," *arXiv preprint arXiv:2410.24164,* 2024.
[50] P. Intelligence, K. Black, N. Brown, J. Darpinian, *et al.*, "$\pi_{0.5}$: a Vision-Language-Action Model with Open-World Generalization," *arXiv preprint arXiv:2504.16054,* 2025.
[51] P. Intelligence, A. Amin, R. Aniceto, A. Balakrishna, *et al.*, "$\pi^*_{0.6}$: a VLA That Learns From Experience," *arXiv preprint arXiv:2511.14759,* 2025.
[52] M. Zawalski, W. Chen, K. Pertsch, O. Mees, C. Finn, and S. Levine, "Robotic Control via Embodied Chain-of-Thought Reasoning," in *8th Annual Conference on Robot Learning,* 2024.
[53] Q. Zhao, Y. Lu, MJ. Kim, *et al.*, "Cot-vla: Visual chain-of-thought reasoning for vision-language-action models," in *Proceedings of the Computer Vision and Pattern Recognition Conference*, 2025, pp. 1702-1713. https://doi.org/10.1109/cvpr52734.2025.00166.
[54] D. Liu, F. Yang, X. Liao, and X. Lyu, "Diablo: A 6-dof wheeled bipedal robot composed entirely of direct-drive joints," in *2024 IEEE/RSJ International Conference on Intelligent Robots and Systems (IROS)*, 2024: IEEE, pp. 3605-3612. https://doi.org/10.1109/iros58592.2024.10801943.